%% file: main.tex
\documentclass[manuscript]{acmart}
\usepackage{booktabs}
\usepackage{multirow}
\usepackage{natbib}
\usepackage{graphicx}
\usepackage{adjustbox} 
\usepackage{longtable}
\usepackage{array}
\usepackage{tikz}
\usepackage{booktabs}  
\usepackage{enumitem}
\usepackage{geometry}  
\usepackage[edges]{forest}
\usetikzlibrary{shadows, calc, positioning}
\definecolor{hidden-draw}{RGB}{20,68,106}
\definecolor{hidden-pink}{RGB}{255,245,247}

\AtBeginDocument{%
  }

\setcopyright{acmlicensed}
\copyrightyear{2025}
\acmYear{2025}
\acmDOI{XXXXXXX.XXXXXXX}

\acmConference[Conference acronym 'XX]{Make sure to enter the correct
  conference title from your rights confirmation emai}{June 03--05,
  2018}{Woodstock, NY}
\acmISBN{978-1-4503-XXXX-X/18/06}

\begin{document}

\title{Evaluating LLM-based Agents for Multi-Turn Conversations: A Survey}

\author{Shengyue Guan}
\affiliation{%
  \institution{Microsoft}
  \country{China}
}
\author{Jindong Wang}
\affiliation{%
  \institution{Microsoft}
  \country{China}
}
\author{Jiang Bian}
\affiliation{%
  \institution{Microsoft}
  \country{China}
}
\author{Bin Zhu}
\affiliation{%
  \institution{Microsoft}
  \country{China}
}
\author{Jian-guang Lou}
\affiliation{%
  \institution{Microsoft}
  \country{China}
}
\author{Haoyi Xiong}
\email{haoyi.xiong.fr@ieee.org}
\affiliation{%
  \institution{Microsoft}
  \country{China}
}

\begin{abstract}
This survey examines evaluation methods for large language model (LLM)-based agents in multi-turn conversational settings. Using a PRISMA-inspired framework, we systematically reviewed nearly 250 scholarly sources, capturing the state of the art from various venues of publication, and establishing a solid foundation for our analysis. Our study offers a structured approach by developing two interrelated taxonomy systems: one that defines \emph{what to evaluate} and another that explains \emph{how to evaluate}. The first taxonomy identifies key components of LLM-based agents for multi-turn conversations and their evaluation dimensions, including task completion, response quality, user experience, memory and context retention, as well as planning and tool integration. These components ensure that the performance of conversational agents is assessed in a holistic and meaningful manner. The second taxonomy system focuses on the evaluation methodologies. It categorizes approaches into annotation-based evaluations, automated metrics, hybrid strategies that combine human assessments with quantitative measures, and self-judging methods utilizing LLMs. This framework not only captures traditional metrics derived from language understanding, such as BLEU and ROUGE scores, but also incorporates advanced techniques that reflect the dynamic, interactive nature of multi-turn dialogues. Together, these frameworks summarize the current status quo, expose limitations in traditional practices, and provide a structured blueprint for improvement. Based on the summarization of existing studies, we identify several challenges and propose future directions, including the development of scalable, real-time evaluation pipelines, enhanced privacy-preserving mechanisms, and robust metrics that capture dynamic multi-turn interactions. Our contributions bridge historical insights with modern practices, paving the way for next-generation, reliably evaluated conversational AI systems and offering a comprehensive guide for researchers and practitioners.
\end{abstract}




\maketitle

\section{Introduction}
Multi-turn conversational agents powered by large language models (LLMs) mark a significant leap in conversational AI~\cite{hassan-graham-2024-advancing}, offering complex, context-aware interactions across multiple turns—essential for applications like customer service and personal assistants~\cite{chen2024personacentricmetamorphicrelationguided, Xi2024ResearchAA}. Evaluating these systems poses unique challenges: methods must be automated, reproducible, and aligned with human judgments to ensure real-world reliability~\cite{yi2024surveyrecentadvancesllmbased}. 

Historically, multi-turn dialogue dates back to ELIZA~\cite{10.1145/365153.365168}, developed in the 1960s, which maintained simple states through pattern matching rather than complex context modeling. Subsequent progress incorporated finite-state machines, frame-based approaches, and statistical methods, culminating in the application of hidden Markov models (HMMs)~\cite{1165342} for spoken language understanding. Deep neural networks and Transformer architectures~\cite{10.5555/3295222.3295349} then revolutionized context modeling, leveraging larger datasets and advanced training methods (including supervised and reinforcement learning) to handle more complex dialogues. In nowadays, multi-turn conversational systems often build on large-scale pre-trained language models, or namely Large Language Models (LLMs)~\cite{OpenAI2023ChatGPT}, enabling advanced reasoning, continuity, and context handling over extended interactions.

\begin{figure}[h]
    \centering
    \includegraphics[width=0.75\textwidth]{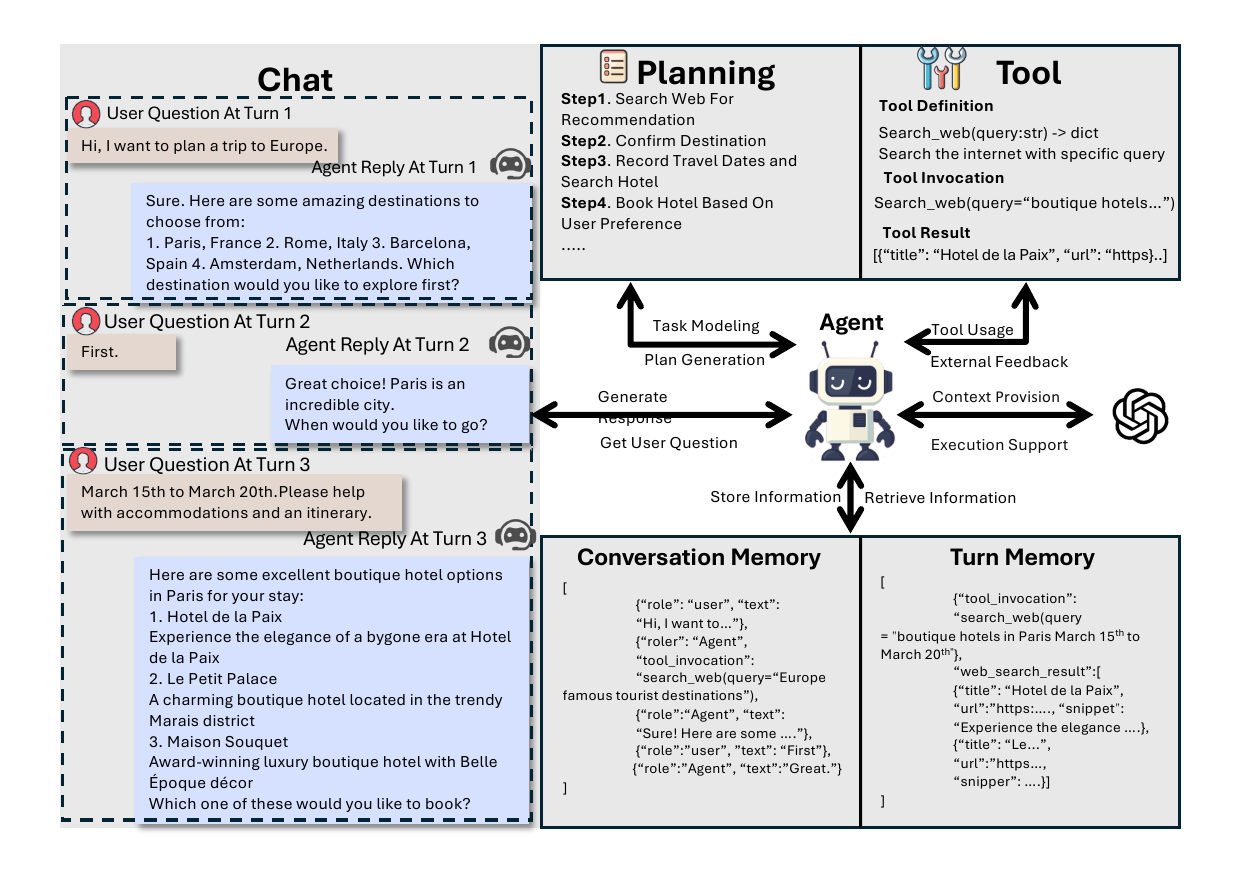}
    \caption{An example of Multi-turn Conversational Agents based on LLMs}
    \label{fig:multi_turn_system}\vspace{-2mm}
\end{figure}

\subsection{LLM-based Multi-turn Conversational Agent}
LLMs, such as ChatGPT~\cite{OpenAI2023ChatGPT}, GPT-4~\cite{OpenAI2023GPT4}, and GPT-o1~\cite{openai2025learning}, are essential for multi-turn conversation, enabling coherent responses over extended interactions. Trained on vast text corpora, LLMs predict subsequent parts of a conversation based on previous turns, and handle complex linguistic structures while maintaining conversational context~\cite{Acharya2023Towards}, essential for generating appropriate responses throughout extended interactions.

An LLM-based multi-turn conversational agent is a chatbot integrate LLM as core with additional three core capabilities—dynamic tool use, persistent memory, and sequential planning—so that it can decompose complex user requests into actionable steps \cite{singh-etal-2024-personal, Jbene2024IntentDF}, call external APIs or services as needed, and remember both recent dialogue and long-term user preferences across turns \cite{kwon-etal-2023-ground, qiu-etal-2024-smile,cho-etal-2023-integrative, Li2020ImproveRD}. Unlike a plain LLM that simply generates text in response to each prompt, an agent autonomously decides when to retrieve or act on external information, tracks context over multiple sessions, and orchestrates a coherent, goal-directed dialogue strategy.

\autoref{fig:multi_turn_system} illustrates an agent-based conversational system. Here, the user plans a trip to Europe; the agent suggests destinations, acknowledges a selection (Paris), confirms dates, and provides accommodation options. The agent’s planning process \cite{huang2024understandingplanningllmagents} breaks the request into actionable steps, narrowing the destination, confirming dates, and searching for relevant hotels. Tool-use \cite{openai_chatgpt_plugins, press2023measuringnarrowingcompositionalitygap, gao2023palprogramaidedlanguagemodels}, such as invoking \texttt{search\_web} for internet queries, yields structured data integrated into the final response. Two types of memory~\cite{Cai2022MemoryGW, liu2024llmconversationalagentmemory}--conversation and turn memory--preserve dialogue context \cite{zhang2024advancingconversationalpsychotherapyintegrating, zhang-etal-2022-history}. The LLM interprets queries, formats responses, and manages complex language tasks \cite{nan-etal-2024-evaluating}, ensuring coherently adapted answers based on user inputs. 

\begin{figure}[h]
    \centering
    \includegraphics[width=0.75\textwidth]{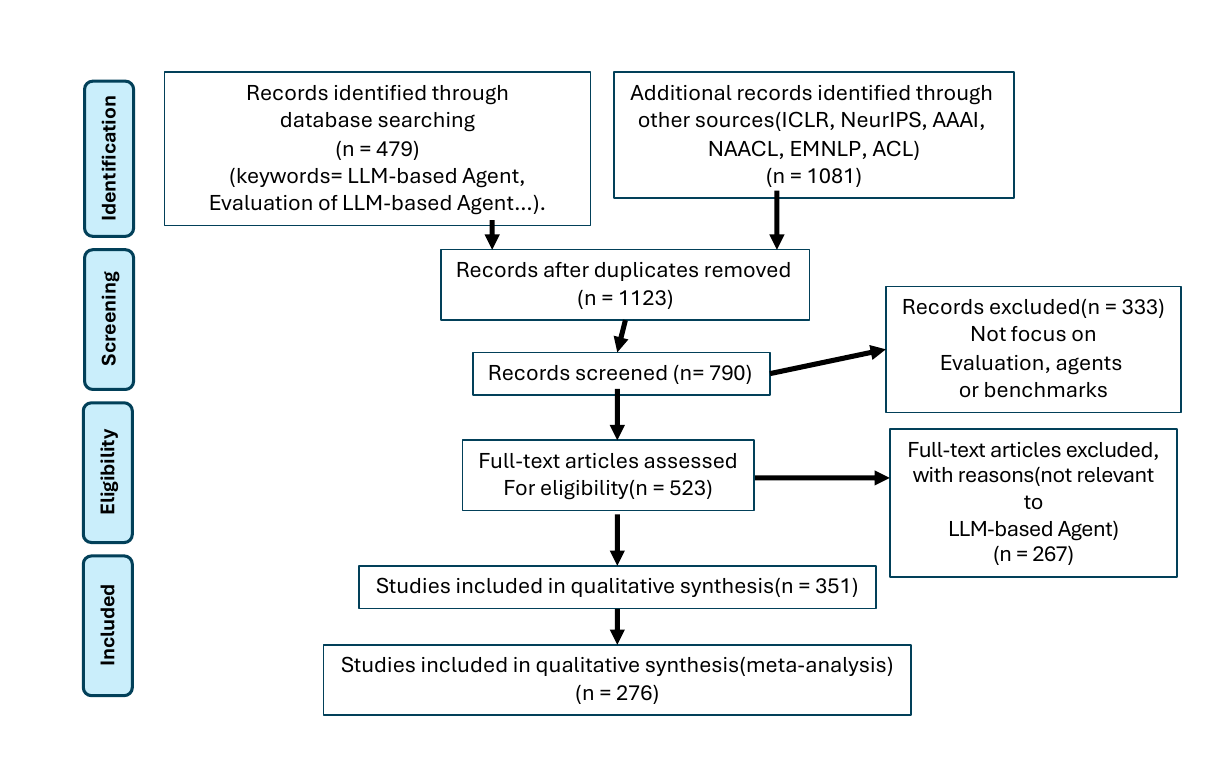}
    \caption{Selection Process for Papers Evaluating LLM-based Agents in Multi-turn Conversations}
    \label{fig:PRISMA}\vspace{-3mm}
\end{figure}

\input{taxonomy}

\subsection{The Scope of This Survey}
This paper surveys evaluation methods for LLM-based multi-turn conversational agents, focusing on (1) \emph{which aspects to evaluate} and (2) \emph{how to evaluate them}. It outlines key targets (task success, response quality, end-to-end user experience) and reviews existing methodologies and datasets, encompassing both automatic metrics and human evaluation protocols. Drawing on literature from 2017 to 2025, this survey follows a methodology inspired by PRISMA~\cite{page2021prisma}, outlined in \autoref{fig:PRISMA}. An initial Google Scholar search retrieved 479 papers\footnote{We use the keywords such as ``evaluation'', ``performance assessment'', ``benchmark'', ``multi-turn conversation/dialogue'', ``multi-round conversation/dialogue'', ``Large Language Model (LLM)'', ``LLM-powered agent'', ``LLM-based agent'', ``LLM agent'' for search.}, supplemented by 1,081 more from venues such as ICLR, NeurIPS, AAAI, NAACL, EMNLP, and ACL (2022–2024). After merging duplicates (1,123 unique records), we excluded 333 irrelevant works, leaving 790 papers; of these, 267 lacked focused coverage of LLM-based agents and were excluded. Ultimately, 351 papers progressed to qualitative synthesis, with 276 offering targeted evaluation insights.

Covering foundational Transformer-based research (2017–2019), the rise of LLMs (2020–2022), and recent multi-turn evaluation advancements (2023–2025), the survey spans approximately 200 papers comprising peer-reviewed conference articles (40\%), journal articles (25\%), preprints (20\%), and industry reports (15\%). Across diverse evaluation data, metrics, and methods, it presents a balanced overview of the rapidly evolving field of the evaluation methodologies for LLM-based multi-turn conversational agents.

\subsection{Our Taxonomy}
Compared to prior research, our taxonomy (a brief representation shown in \autoref{fig:taxonomy}) reviews the literature that studies the evaluation of LLM-based multi-turn conversational agents by systematically addressing three key dimensions:
\begin{itemize}
     \item \textbf{What to Evaluate?} Comprehensive Evaluation Goals: Defining broad objectives to assess agent capabilities across various tasks and interactions. This includes evaluating aspects such as task success rates, response quality, user engagement, and overall user experience.
     
     \item \textbf{How to Evaluate?} Diverse Methodologies: Exploring a range of evaluation techniques, including data-driven annotations, automatic evaluation metrics, and innovative metric designs. This dimension emphasizes the integration of both human and machine-based assessment tools to provide a holistic evaluation framework.
    \end{itemize}
Our taxonomy categorizes evaluation methods based on these dimensions, providing a structured approach to understanding and improving the assessment of multi-turn conversational agents. By systematically addressing what aspects to evaluate and how to evaluate them, our taxonomy lays the foundation for more reliable and comprehensive evaluation practices. Additionally, by identifying future challenges, it further guides researchers and practitioners in developing next-generation evaluation frameworks that can keep pace with advancements in conversational AI.

\subsection{Comparisons with Other Existing Surveys}
Existing surveys in the related areas have distinct focuses but often leave critical gaps. For example, \citet{arora2013dialoguesystembriefreview} give only a brief overview of dialogue systems, while classifications into task-oriented vs. non-task-oriented models can overlook agent–system interactions \cite{10.1145/3166054.3166058}. Recent surveys \cite{ni2022recentadvancesdeeplearning, yi2024surveyrecentadvancesllmbased,10.1007/s00521-023-09322-1,hu2024surveylargelanguagemodelbased,Wang_2024} each tackle different perspectives but do not address interactions between agentic components in detail or the unique challenges of multi-turn evaluations. \citet{li2024reviewprominentparadigmsllmbased} discuss LLM-based agent capabilities without focusing on multi-turn dialogue, and \citet{10.1145/3641289} review broader LLM evaluation dimensions without delving into multi-turn specifics. \citet{DBLP:journals/corr/abs-2501-09959} discuss evaluators, such as GPT-4 for assessing multi-turn interactions. However, it lacks detailed descriptions of both annotation-based and annotation-free evaluation methods.

Our taxonomy fills these gaps by systematically exploring what aspects to evaluate and how. We define comprehensive goals, discuss data-driven annotation and automatic evaluation, and study robust metrics in multi-turn evaluation. Unlike earlier work \cite{arora2013dialoguesystembriefreview,10.1145/3166054.3166058,ni2022recentadvancesdeeplearning,yi2024surveyrecentadvancesllmbased}, our framework addresses future challenges (e.g., scalability, real-world applicability) and offers a structured, holistic approach to assessing LLM-based multi-turn conversational agents. This perspective is vital for building reliable, adaptable dialogue systems capable of handling complex multi-turn interactions effectively.

\section{What to evaluate? The evaluation goals and target}
Evaluating LLM-based multi-turn conversational agents is essential in an era where these systems power interactive applications like virtual assistants, customer support bots, and collaborative tools. Unlike single-turn models, multi-turn agents must handle ongoing dialogues, maintaining context across interactions, adapting to user inputs, and executing complex tasks.This requires assessing not just isolated responses but the interplay of various components that drive overall performance, reliability, and user satisfaction. However, evaluations can be fragmented without a structured approach. To address this, we categorize evaluation goals into four interconnected areas, following a logical workflow: starting with the end-to-end experience to capture real-world usability, then drilling down into specific functionalities like actions/tools, memory, and planning \cite{Reimann2023PredictingIQ, Hendrycks2020MeasuringMM, gritta2024humanrankevalautomaticevaluationlms}.

\subsection{Evaluating the End-to-end Experience of LLM-based Agents For Multi-Turn Conversation}
It is natural to prioritize evaluating the end-to-end experience in the first place, as this holistic approach directly captures real-world performance before delving into granular metrics. Multi-turn dialogues inherently differ from isolated single-turn queries in that they often encompass multiple sequential or interleaved tasks within a single conversation. This complexity necessitates evaluating an agent’s performance at the dialogue level, beyond individual exchanges. Specifically, we must assess not only whether each task is completed but also how well tasks are executed. In other words, even if an agent successfully fulfills each requested task, we still examine the overall quality of task completion and the interaction process: does the agent transition smoothly between tasks, maintain context and coherence, and deliver responses that keep the user satisfied and engaged without producing harmful or irrelevant content? For example, an agent may correctly answer each question (achieving high task completion) but still fail overall if its responses are confusing, inconsistent, or inappropriate. Hence, We present an overview of End-to-end Experience evaluation goals for LLM-based agents in multi-turn interactions. The representative works could be categorized into five dimensions: (1) \emph{Task Completion in Multi-Turn Conversations}, measuring how well agents fulfill user requests using metrics like completion rate; (2) \emph{Multitask Capabilities}, assessing expertise across domains such as mathematics, coding and etc; (3) \emph{Interaction Patterns}, analyzing structures like recollection, follow-up, and expansion to assess dialogue coherence and (4) \emph{User Experience and Safety}, addressing satisfaction, engagement, and safeguards against harmful content, including adversarial attacks and prompt leakage.

\subsubsection{Task Completion in Multi-Turn Conversations}
In multi-turn conversations, task completion extends beyond simple query resolution to encompass specific considerations such as maintaining contextual coherence, mitigating error propagation across turns, and adapting to evolving user intents.These factors introduce unique challenges, including the risk of misalignment due to ambiguous references or incomplete information, necessitating robust evaluation metrics that capture not just final outcomes but the iterative process of goal achievement.\citet{Reimann2023PredictingIQ} introduced objective metrics for assessing interaction quality through task completion rates and unrecognized utterance tracking, addressing limitations in existing evaluation approaches with a three-level scoring system that examines usability, likability, conversation quality, and interaction metrics. This framework was validated using Cookpanion, a recipe recommendation agent employing twelve conversation patterns. Additionally, \citet{Deriu_2020} emphasizes measuring task completion through user feedback and system logs, determining success based on request fulfillment within conversations. However, both studies lack detailed analysis of failure causes, such as miscommunication or system limitations, leaving a critical gap in understanding multi-turn interaction challenges while highlighting the need for improved evaluation frameworks. Furthermore, language models frequently serve varied user needs beyond single-use scenarios. \citet{Hendrycks2020MeasuringMM} introduce a comprehensive benchmark evaluating multitask capabilities across 57 distinct tasks in fields like mathematics and law. Task completion is measured by the accuracy of model responses to domain-specific questions. Moreover, \citet{gritta2024humanrankevalautomaticevaluationlms} propose an evaluation framework for conversational agents that includes multiple expert-rated answers for each question, enhancing the understanding of how models manage multi-turn interactions and ultimately improving user experience.

Although current frameworks effectively quantify task completion and multitask performance, they generally lack detailed failure-mode analysis and user-centric feedback loops. Integrating error diagnostics and richer interaction metrics will be essential for developing more robust evaluation methods for conversational agents.

\subsubsection{Interaction Patterns}
In multi-turn conversations, interaction patterns show the flow of dialogue, incorporating specific considerations such as pattern recognition for contextual cues, seamless transitions between engagement modes, and alignment with user-driven evolutions to sustain coherence and depth.These considerations highlight the need for agents to dynamically interpret and respond to evolving exchanges, beyond static responses, thereby fostering more natural and effective interactions. 
\citet{Zheng2023JudgingLW} developed MT-Bench, a multi-turn benchmark evaluating eight categories (writing, roleplay, extraction, reasoning, mathematics, coding, STEM, and humanities) using LLMs as evaluators. Extending this work with a million-conversation dataset across 25 LLMs \cite{zheng2024lmsyschat1mlargescalerealworldllm}, \citet{kwan2024mtevalmultiturncapabilitiesevaluation} identified four key interaction patterns: recollection, expansion, refinement, and follow-up. \citet{Bai_2024} further contributed a three-tier taxonomy (Perceptivity, Adaptability, Interactivity) analyzing 4,208 turns across 1,388 dialogues.

These interaction patterns demonstrate how LLMs progressively build context, enrich content, refine responses, and guide conversations in multi-turn dialogues. The three-tier taxonomy further clarifies their comprehensive capabilities and adaptive strategies across diverse tasks and contexts.

\subsubsection{User Experience and Safety}
Evaluating user experience and safety in multi-turn conversations demands specific considerations, such as keeping users engaged through flexible feedback that matches their changing intention, and strengthening protections against risks that grow worse in stacked contexts—such as mix-ups in references or ongoing attacks. These points matter because drawn-out talks can turn small hiccups into big drop-offs or spread quiet dangers into chain reactions, pushing for evaluation methods that track the full flow of interactions instead of just quick snapshots.User experience evaluation in multi-turn agents \cite{miehling-etal-2024-language} has evolved with \citet{xu-etal-2022-endex}'s human-reaction model for measuring dialogue engagement, while \citet{liu2024convbenchmultiturnconversationevaluation} expanded this to LVLMs through a three-level capability hierarchy. Safety concerns \cite{tong-etal-2024-securing,bassani-sanchez-2024-guardbench} are addressed by the CoSafe dataset proposed by \citet{yu2024cosafeevaluatinglargelanguage}, revealing vulnerabilities to coreference-based attacks. Additionally, prompt leakage threats \cite{agarwal-etal-2024-prompt} highlight the need for robust privacy and intellectual property protection.
\citet{dong-etal-2024-attacks} explore the impact of backdoor attacks on multi-turn conversational agents, which could result in harmful, biased, or misleading outputs. The work by \citet{mehrabi-etal-2022-robust} demonstrates that agents, when exposed to imperceptible toxic triggers, can produce content that is unintentionally offensive or harmful. These findings underscore the critical importance of comprehensive safety evaluation for LLM-based agents in multi-turn conversations, particularly when deployed in high-stakes domains such as medicine or education where such evaluations become essential for responsible implementation. A comprehensive safety evaluation should systematically test a multi-turn agent’s resilience against coreference attacks, prompt leakage, backdoor insertion, and stealthy toxic triggers, simulating each threat scenario to validate robust, harm-free performance.

\subsection{Evaluating The Action and Tool-use Components of LLM-based Agents for Multi-turn Conversation}
The evaluation of action and tool-use components must address a hierarchy of challenges: from the fundamental capability to execute commands, to the complexity of reasoning across turns, and finally to the trustworthiness of the outcomes. In multi-turn dialogues, agents act as bridges between user intent and external execution. This role necessitates a three-layered evaluation framework. First, the agent must demonstrate basic proficiency in identifying and executing discrete external commands, serving as a functional interface \cite{openai_chatgpt_plugins, shi-etal-2024-learning}. Second, beyond isolated actions, the agent must sustain a coherent "chain of execution" where outputs from one turn inform the inputs of the next, requiring sophisticated multi-step reasoning \cite{gu-etal-2024-middleware, qin2023toolllmfacilitatinglargelanguage, gao2023palprogramaidedlanguagemodels, press2023measuringnarrowingcompositionalitygap}. Finally, as agents gain the power to act, the risk of "action hallucination"—invoking non-existent tools or misinterpreting outputs—grows, making reliability a critical safety constraint \cite{zhang2024toolbehonestmultilevelhallucinationdiagnostic,chen-etal-2024-diahalu}.

Based on this hierarchical necessity, representative works are categorized into three dimensions: (1) \emph{API Interaction and Dynamic Tool-Use}, which evaluates agents' ability to adapt to different tools and perform tasks without explicit instructions; (2) \emph{Multi-step Tool Selection and Reasoning}, which includes frameworks and datasets for assessing LLM-based agent ability to select and apply multiple tools across sequential, multi-turn interactions; and (3) \emph{Reliability and Hallucination in Tool-Use}, which assesses accuracy and examines scenarios of incorrect outputs due to tool response misinterpretation.

\subsubsection{API Interaction and Dynamic Tool-Use}
Evaluating API interaction and dynamic tool-use requires a dual focus on interaction precision and adaptive orchestration. Specifically, the evaluation must verify whether an agent can accurately map user intents to specific API calls while also assessing its capacity to autonomously plan and chain tools for complex, unseen tasks without rigid instructions.
In this context, \citet{xu2023toolmanipulationcapabilityopensource} evaluate LLMs' performance in intent recognition, response selection, and tool utilization, highlighting their ability to access real-time data and maintain context across conversation turns, though with varying proficiency. \citet{shen2023hugginggptsolvingaitasks} showcase ChatGPT's capacity for task planning by integrating multiple AI models across diverse domains to enhance performance. \citet{patil2023gorillalargelanguagemodel} improve API interaction accuracy and adaptability by incorporating Torch and TensorFlow APIs alongside a real-time document retriever, enabling precise and dynamic tool usage. HuggingGPT \cite{shen2023hugginggptsolvingaitasks} further demonstrates the effective orchestration of multiple AI models for handling complex tasks. \citet{DBLP:conf/iclr/MialonF0LS24} propose benchmark called GAIA that test agent's proficiency in tool use by presenting question on real world that require long sequences action and API interaction.To summarize, those studies show that API interaction and dynamic tool-use can be evaluated from multiple perspectives, including intent recognition and response selection accuracy; context preservation and consistent tool invocation; real-time data integration and precise document retrieval and long-sequence API chaining proficiency and accuracy.

\subsubsection{Multi-step Tool Selection and Reasoning }
Multi-step tool use requires the agent's "chain of thought" in action. It is not enough to select the right tool once; the agent must demonstrate contextual awareness to know when tools are needed, sequential logic to order them correctly across turns, and domain resilience to maintain this logic in specialized fields like math or coding.
MetaTool \cite{huang2024metatoolbenchmarklargelanguage} evaluates LLMs' tool awareness and decision-making using the ToolE dataset, which tests single- and multi-tool scenarios. It highlights challenges in tool selection, revealing gaps in LLM capabilities. \citet{wang2024mtubenchmultigranularitytoolusebenchmark} introduced MTU-Bench to assess multi-tool usage and multi-step interactions across scenarios, including single- and multi-turn tasks, enhancing action execution in conversations. \citet{nijkamp2023codegen} proposed a program synthesis benchmark focusing on breaking down complex programming tasks, while \citet{wang2024mintevaluatingllmsmultiturn} developed a Python-based benchmark for evaluating tool-use in responses. Mathematical reasoning benchmarks were introduced by \citet{kurtic-etal-2024-mathador} and \citet{sun-etal-2024-mm}, focusing on dialogue contexts. \citet{guo-etal-2024-ctooleval} designed a benchmark for assessing multi-turn tool-use in Chinese societal applications, and \citet{zhuang2023toolqadatasetllmquestion} emphasized open-ended tool usage in question answering. Lastly, \citet{huang-etal-2024-planning-creation} proposed a comprehensive benchmark to evaluate the full tool-use process, including planning and creation in complex tasks. These studies suggest that a unified evaluation framework for multi-step tool selection and reasoning should assess invocation precision, operational efficiency, and error recovery. It should also evaluate tool awareness, planning capability, execution accuracy, and multi-turn coherence.

\subsubsection{Reliability and Hallucination in Tool-Use}
In tool-use scenarios, hallucinations are not merely textual inaccuracies but functional failures that can trigger unintended actions. Therefore, reliability evaluation must transcend simple error counting to establish a safety perimeter, requiring protocols that simultaneously diagnose the root causes of failure and enforce strict confidence thresholds before action execution \cite{liu-etal-2024-uncertainty, muthusamy-etal-2023-towards}. 

\citet{zhang2024toolbehonestmultilevelhallucinationdiagnostic} introduce a benchmark addressing hallucinations through depth, involving multi-level diagnostics like solvability detection and missing-tool analysis, and breadth, focusing on toolset limitations. \citet{cao-2024-learn} propose reducing hallucinations at the reasoning stage using soft and hard evaluation methods. In the hard approach, similarity scores between retrieved knowledge and confidence levels are calculated; responses are only provided if the threshold is met, ensuring precision. These studies showed us when evaluating the reliability and hallucination in tool-use, it is essential to comprehensively assess tool limitations, model calibration during reasoning, and multi-level diagnosis of hallucination.

\input{memory_taxonomy}

\subsection{Evaluating the Memory of LLM-based Agents For Multi-turn Conversation}
Memory in agent is not merely a storage unit but a dynamic cognitive resource that bridges the gap between transient interactions and enduring persona continuity. Evaluating this capability requires a dual perspective: a temporal hierarchy that dictates how long information persists, and a representational strategy that determines how efficiently it is accessed.

In multi-turn conversations, the agent faces a constant trade-off: retaining everything creates cognitive overload, while discarding too much breaks continuity. The span of memory resolves this by categorizing information urgency—from immediate turn-level buffers to permanent knowledge stores \cite{Hatalis2024MemoryMT}. Simultaneously, the form of memory addresses the retrieval bottleneck; evaluating whether data should be kept as explicit, searchable text or internalized into the model's parameters directly impacts the agent’s adaptability and inference speed \cite{liu2023agentbenchevaluatingllmsagents, yi2024surveyrecentadvancesllmbased}.

Representative studies can be classified into two dimensions, as illustrated in \autoref{fig:memory taxonomy}: (1) \emph{Memory Span}, which examines the temporal scope of memory, spanning individual turns, entire conversations, and long-term persistence; and (2) \emph{Memory Forms}, which delves into the representations and implementation of memory.

\subsubsection{Memory Spans}
Memory in LLM-based agents operates across a temporal hierarchy that mirrors human cognitive architecture: from \textbf{immediate awareness} (turn memory) that anchors the agent within a single exchange, to \textbf{sustained context} (conversational memory) that weaves coherence across extended dialogues, and finally to \textbf{persistent knowledge} (permanent memory) that transcends individual sessions to support long-term personalization. Effective evaluation of these memory spans demands distinct but complementary criteria: \emph{turn}, \emph{conversation}, and \emph{permanent} memory. Turn memory captures details within a single dialogue turn; for instance, in \autoref{fig:multi_turn_system}, when a user requests boutique hotels in Paris, the agent’s action (SearchWeb) and its results are recorded, enabling appropriate responses based on the immediate context \cite{Wang2020DualDM}. \citet{Shen2022KWickChatAM} emphasizes the importance of context and dialogue history for coherence, while \citet{Cai2022MemoryGW} introduces a sensory memory module that converts the current utterance into word- and sentence-level embeddings.These studies showed that effective evaluation of turn memory requires emphasize dynamic and accurate per-turn updates, efficient retrieval and management of stored context, and tangible improvements in task performance within each dialogue turn.

Meanwhile, conversational memory extends this retention across multiple turns by tracking user preferences and details throughout extended dialogues, with benchmarks like LongEval \citep{leng2024longcontextragperformance} and SocialBench \citep{chen-etal-2024-socialbench} assessing retention over 40+ utterances, and \citet{maharana2024evaluatinglongtermconversationalmemory} demonstrating dialogues spanning 600 turns and 16K tokens by incorporating question answering for factual recall, event summarization for temporal reasoning, and multimodal generation for persona consistency. In addition, \citet{zhang2024memsimbayesiansimulatorevaluating} developed QA pairs from simulated conversations and \citet{liu2024llmconversationalagentmemory}'s RAISE architecture enhances ReAct with dual memory components for improved context tracking. \citet{xu2021goldfishmemorylongtermopendomain} shed light on long-term conversations, examining scenarios where interlocutors converse intermittently over multiple sessions, gradually accumulating shared context. \citet{jang2023conversation} extend the timeframe from the initial few days to intervals ranging from hours to years, illustrating how memory retention can scale over significantly varied time spans. Similarly, \citet{maharana2024evaluatinglongtermconversationalmemory} target conversations lasting multiple months, filling an essential gap between short-term interactions and long-range, multi-year studies. To summarize, Effective evaluation of conversational memory requires focusing on long-term context retention, integration for consistent and coherent responses, and adaptability in memory retrieval. Benchmarks that test extended dialogues, factual recall, and temporal reasoning over varied timeframes are essential to ensure robust and scalable conversational performance.

Permanent memory enables the agent to retain critical information across conversations, fundamentally supporting long-term interactions; for example, \cite{castillobolado2024promptsdynamicconversationalbenchmarking} showcases an agent storing a user’s preference for boutique hotels near historic landmarks to manage concurrent tasks and context switches, while \citet{huang2023memorysandboxtransparentinteractive} introduced Memory Sandbox for interactive, user-controlled memory management, and \citet{fu-etal-2024-msi} propose a three-part pipeline—experience selector, insight generator, and insight selector—to extract task-relevant memory. These studies showed us that evaluating memory spans in LLM agents requires assessing how effectively they retain and utilize information across turn, conversational, and permanent memory. Key aspects include memory capacity, retention over extended dialogues, cross-session consistency, and the ability to recall and apply facts, preferences, and events under different scenarios.

\begin{figure}[h]
    \centering
    \includegraphics[width=0.9\textwidth]{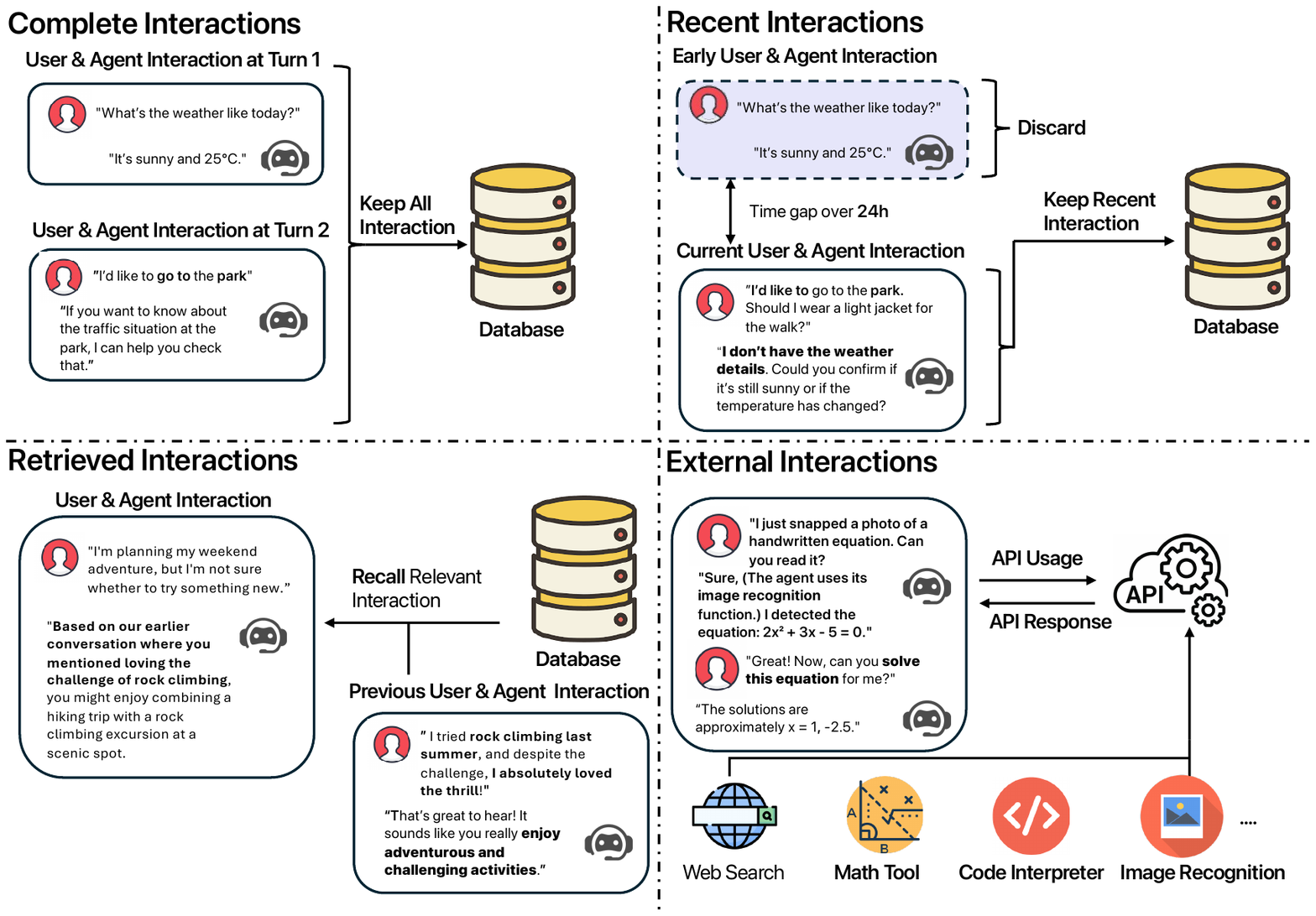}
    \caption{Four Types of User-Agent Interactions \cite{zhang2024surveymemorymechanismlarge}: \textbf{Complete Interactions}: This refers to the database storing all interactions between the user and the agent. Every conversation or action is captured and retained for future reference; \textbf{Recent Interactions}:In this case, if the user asks for weather details and the time gap exceeds 24 hours, the information is discarded.; \textbf{Retrieved Interactions}: This occurs when the agent recalls past interactions.\textbf{External Interactions}: In this scenario, the user uploads an image, and the agent uses image recognition to analyze the content. The agent might then use a math tool to solve an equation that is written on the image.
    }
    \label{fig:memory_form}
\end{figure}

\subsubsection{Memory Forms}
The architectural choice of memory representation fundamentally determines an agent's capacity for context retention, retrieval efficiency, and adaptability across diverse interaction scenarios.
It can be broadly categorized into \emph{textual} and \emph{parametric forms} \cite{zhang2024surveymemorymechanismlarge}. \emph{Textual forms} include \emph{complete interaction}—which retains the entire dialogue history for complex planning tasks and extended reasoning (illustrated in \autoref{fig:memory_form}), with examples in \cite{zhang2023historyawarehierarchicaltransformermultisession, wang2023fccfusingconversationhistory, huang2024doesconversationlengthimpact} and further evaluated by \citet{li2023how} and \citet{maharana2024evaluatinglongtermconversationalmemory} via the LoCoMo dataset and evaluation frameworks in \cite{lei2024s3evalsyntheticscalablesystematic, pal2023giraffeadventuresexpandingcontext, tworkowski2023focusedtransformercontrastivetraining, bae2022updatedmemorymanagementlongterm}), \emph{recent interaction}—which preserves only the most immediate exchanges to balance context retention with computational efficiency (as implemented in the SCM system of \cite{Liang2023UnleashingII}, the TiM framework in \cite{Hoxha2024BlockchainAA}, DiagGPT \cite{cao2024diaggptllmbasedmultiagentdialogue}, and discussed in \cite{wang2024userbehaviorsimulationlarge, DBLP:journals/corr/abs-2501-17399}), \emph{retrieved interaction}—which selectively recalls dialogue segments based on semantic relevance, contextual significance, and thematic coherence (using methods such as cosine similarity in \cite{sarch-etal-2023-open, park2023generativeagentsinteractivesimulacra}, semantic similarity in \cite{ong2024lifelongdialogueagentsrelationaware}, and enhanced by frameworks like MemoryBank \cite{zhong2023memorybankenhancinglargelanguage} with FAISS \cite{johnson2017billionscalesimilaritysearchgpus}, thought-based retrieval in \cite{liu2023thinkinmemoryrecallingpostthinkingenable}, RecMind \cite{wang-etal-2024-recmind}, and RET-LLM \cite{modarressi2024retllmgeneralreadwritememory}), and \emph{external interaction}—which integrates external tools and APIs for dynamic data retrieval (as shown in \cite{xu2023toolmanipulationcapabilityopensource, qin2023toolllmfacilitatinglargelanguage}).These studies showed that evaluating textual memory demands examining coverage of full-history retention, context completeness versus efficiency for recent windows, precision and recall in selective retrieval, and correctness and latency when integrating external data. In contrast, \emph{parametric forms} encode memory directly within model parameters, thereby circumventing context length constraints; this category comprises \textbf{fine-tuning approaches}—demonstrated by \citet{zhang2024personalizedllmresponsegeneration}, \cite{li2022largelanguagemodelscontrollable}, \citet{tan2024democratizinglargelanguagemodels}, and \cite{fountas2024humanlikeepisodicmemoryinfinite}—and \textbf{memory editing techniques} that allow targeted modification of parameters, as proposed by \cite{mitchell2022memorybasedmodeleditingscale}, \cite{qiao-etal-2024-comem}, \cite{zeng2024famefactualmultitaskmodel}, \cite{tack2024onlineadaptationlanguagemodels}, and \cite{mao2024editingpersonalitylargelanguage} (also discussed in \cite{zhang2024surveymemorymechanismlarge}). Collectively, these diverse memory forms enable agents to maintain coherence, personalize responses, and enhance overall performance in multi-turn conversations. These studies demonstrated us that evaluating memory forms includes the capacity and retrieval accuracy of different textual forms (full-history, recent, retrieved, external), the adaptability and updating control of parametric memory, and the trade-offs between efficiency, generalization, and scalability.

\input{planner_taxonomy}

\subsection{Evaluating the Planner of LLM-based Agents For Multi-turn Conversation}
To rigorously evaluate the planner in multi-turn conversations, we view the planning process not as a linear pipeline, but as a continuous control loop comprising four essential dimensions (see \autoref{fig:planner taxonomy}): First, the agent must establish a grounded understanding of the environment and objectives, acting as the foundational map for interaction. Second, to bridge the gap between high-level intent and low-level execution, the agent must possess the architectural ability to dismantle complex goals into manageable sequences. Third, as conversations are inherently non-stationary, the planner requires the elasticity to navigate shifting user intents and environmental feedback in real-time \cite{huang2024understandingplanningllmagents}. Finally, to ensure reliability and safety, the system must employ metacognitive mechanisms to validate paths and select optimal strategies before and during execution. This framework organizes our analysis of the planning pipeline, detailed as follows:
\emph{Task Modeling}, which examines both \emph{Task Representation} (defining task conceptualization \cite{huang2024understandingplanningllmagents}) and \emph{Context Modeling} (capturing situational awareness for planning); \emph{Task Decomposition}, which analyzes strategies for breaking complex tasks into executable sub-components \cite{zhang2024probingmultiturnplanningcapabilities}; \emph{Adaptation and Control}, which assesses dynamic response capabilities during multi-turn interactions \cite{muise2019planninggoalorienteddialoguesystems}; and \emph{Reflection}, which evaluates verification and selection mechanisms through \emph{Plan Verification} (checking execution feasibility)\cite{zhang2024probingmultiturnplanningcapabilities} and \emph{Plan Selection} via \emph{In-Generation} filtering or \emph{Post-Generation} evaluation.

\subsubsection{Task Modeling}
Task modeling lays the groundwork, but specific considerations—like framing tasks with sharp clarity to lock in goals and anchoring context to track evolving details—are essential to avoid early mix-ups that ripple into full breakdowns.These steps matter because fuzzy starts or lost threads can derail the whole conversation, demanding evaluations that check not just what’s defined but how it holds up over turns. This effort involves two fundamental aspects: \emph{Task Representation} and \emph{Context Modeling}. {Task Representation} encompasses the clear delineation of activities the agent must execute, including objectives, actions, and expected outcomes, thereby empowering effective multi-turn conversations. For example, \citet{2024multi-agent} presents MA-LAMA, a multi-agent temporal planner that uses a factored, centralized approach to translate temporal tasks into constrained snap-actions, while \cite{singh2024twostepmultiagenttaskplanning} integrates classical planners with LLMs to improve goal decomposition for two agents, and LaMMA-P \cite{zhang2024lammapgeneralizablemultiagentlonghorizon} utilizes a Language Model-Driven Multi-Agent PDDL Planner for effective task management. Additionally, a Planner-Reasoner framework for multi-task reasoning agents is introduced in \citet{lyu2022primaplannerreasonerinsidemultitask}. In parallel, {Context Modeling} pertains to the agent's ability to preserve and leverage contextual information throughout interactions, ensuring responses remain coherent and contextually appropriate. Proficient context modeling monitors antecedent interactions, user preferences, and influential details, as demonstrated in \cite{feng-etal-2022-represent}. The MNDB model \cite{10.1145/3442381.3449902} employs a ternary-grounding network to simulate natural conversations, while an optimization strategy using self-contained distractions is proposed in \cite{xing2022evaluatingimprovingcontextattention} for enhanced performance. Moreover, \citet{xiao-etal-2024-flowbench} formalizes workflow knowledge to mitigate context hallucinations, and a planner predicting continuous action vectors based on the last-token embedding is detailed in \cite{li2024dialogueactiontokenssteering}. Structural models that integrate state recognizers and structure planners \cite{DBLP:conf/kdd/FuZ023}, as well as the Meta Decision Transformer (Meta-DT) for offline meta-reinforcement learning \cite{wang2024metadtofflinemetarlconditional}, further bolster context modeling capabilities. Previous studies showed us that evaluating task modeling in LLM agents requires rigorous assessment of how tasks are represented and how context is modeled. Key aspects include clarity in structured action planning and dynamic adaptation to evolving objectives and constraints. Robust evaluation should track multi-turn coherence, context utilization, and success rates across diverse scenarios to reveal the agent’s ability to plan, execute, and maintain context throughout complex tasks.

\subsubsection{Task Decomposition}
Task decomposition is pivotal for manageable execution in complex environments, requiring planners to balance three critical dimensions: granularity (the resolution of subtasks), interdependency (the logical coupling between actions), and dynamism (the ability to restructure plans on-the-fly). Earlier approaches focused on static breakdowns, such as 3D structure decomposition \cite{srinivasan2023multiagentcollectiveconstructionusing} and prompt-based planning strategies that outline sequential steps \cite{shen2023hugginggptsolvingaitasks,wang2023planandsolvepromptingimprovingzeroshot, singh2022progpromptgeneratingsituatedrobot}. More recent recursive methods like ADaPT adjust granularity based on LLM capabilities \cite{prasad2024adaptasneededdecompositionplanning}, while MLDT employs multi-level decomposition specifically for spatial planning \cite{wu2024mldtmultileveldecompositioncomplex}.
To address interdependency, neuro-symbolic methods such as LLM-DP combine traditional planning with LLMs to maintain logical coherence \cite{dagan2023dynamicplanningllm,kwon2024fastaccuratetaskplanning}, and TDAG employs a graph-based framework to manage complex subagent interactions \cite{wang2024tdagmultiagentframeworkbased}. 
Finally, for dynamism, PoT offloads computations to external interpreters to reduce burden \cite{chen2023programthoughtspromptingdisentangling}, whereas D-PoT actively adjusts decomposition trees via environmental feedback \cite{zhang2024dynamicplanningllmbasedgraphical}.
Rigorous evaluation must therefore measure these specific aspects: adaptive decomposition strategies, the precision of subtask division, coordination mechanisms, and feedback utilization, ensuring agents can effectively decompose and manage complex tasks in real-world settings.

\subsubsection{Adaptation and Control}
As these agents engage in prolonged interactions, they must demonstrate a robust capacity to recalibrate their responses based on evolving contextual cues and user feedback. It requires a deep understanding of the interplay between environmental dynamics and user-specific goals, enabling agents to maintain relevance and coherence throughout prolonged interactions. We classify existing research based on the environment in which the agent operates. For non-stationary environments, the MARL-CD approach \cite{9864802} detects context changes and adjusts planning strategies, while the action graph planner \cite{10054388} captures state transitions for strategy updates. User intention adaptation employs reasoning path visualization \cite{hao2023intentdialintentgraphbased} and epistemic planning frameworks \cite{Huang_2017} that generate action trees from sensor data. The LARA framework \cite{liu-etal-2024-lara} enhances intent classification through conversation history analysis. \citet{muise2019planninggoalorienteddialoguesystems} introduces declarative agents using planning technology to dynamically adapt dialogues, eliminating specified dialogue trees manually.To Effectively evaluating of adaptation and control centers on their ability to sense shifting environments and user goals, respond with dynamic re-planning, and maintain coherence. Effective assessment should measure interactive adjustment strategies and goal alignment under uncertainty or context change to ensure planners remain relevant and coherent in evolving scenarios.

\subsubsection{Reflection}
Reflection serves as the metacognitive safeguard in the planning pipeline, ensuring that agents do not merely generate actions but rigorously evaluate their validity and optimality before and during execution.This dimension operates through two complementary mechanisms: \emph{Plan Verification} and \emph{Plan Selection}. Specifically, Plan Verification assesses the feasibility and potential outcomes of proposed actions to ensure alignment with the agent’s goals and constraints, thereby identifying pitfalls before execution. Innovative approaches support this process: React \cite{yao2022react} enables guided reasoning with real-time validation; strategic planning and inductive reasoning mechanisms help narrow possibilities \cite{zhang2024probingmultiturnplanningcapabilities}; and model checking techniques address critical issues like deadlocks and safety violations in multi-agent systems \cite{10.1145/3287921.3287947}. Further, frameworks such as Critic-CoT \cite{zheng2024criticcotboostingreasoningabilities} and the Self-Refine process \cite{madaan2023selfrefineiterativerefinementselffeedback} facilitate iterative refinement, while graph-based methods \cite{cao-2024-graphreason} merge similar reasoning steps to form a comprehensive representation of the problem-solving landscape. In parallel, Plan Selection is broadly divided into \textbf{in-generation selection} and \textbf{post-generation selection}. In-generation selection involves choosing the optimal plan during solution formulation; for example, MultiESC \cite{cheng-etal-2022-improving} employs a lookahead strategy for emotional support dialogues, MA-LAMA \cite{2024multi-agent} segments tasks into independent search phases, and techniques using Dialogue Action Tokens (DAT) \cite{li2024dialogueactiontokenssteering} alongside Direct Multi-turn Preference Optimization (DMPO) \cite{shi2024directmultiturnpreferenceoptimization} enhance goal-directed planning. 

Additionally, the Dynamic LLM-Agent Network (DyLAN) \cite{liu2024dynamicllmpoweredagentnetwork} enables dynamic inference-time agent selection with early-stopping to optimize responsiveness, while approaches like AdaPlanner \cite{sun2023adaplanneradaptiveplanningfeedback}, PLANSEARCH \cite{wang2024planningnaturallanguageimproves}, and Plan, Generate and Match (PGM) \cite{10.1007/978-3-031-48421-6_7} further improve candidate plan evaluation. In contrast, post-generation selection leverages structured reasoning frameworks such as Tree of Thoughts (ToT) \cite{yao2023treethoughtsdeliberateproblem} and Graph of Thoughts (GoT) \cite{Besta_2024} to facilitate multi-path exploration and self-reflection, while human-in-the-loop systems like LLM A* \cite{xiao2024llmahumanloop} and dialogue systems generating global paths \cite{liu-etal-2023-mtgp} enhance transparency, and reinforcement learning techniques optimize action space relevance \cite{he2016deepreinforcementlearningnatural}. These integrated mechanisms collectively empower LLM-based agents to refine their decision-making capabilities, adapt to dynamic contexts, and improve overall performance in complex scenarios \cite{shen-etal-2024-smartcal}. 

To summarize, Reflection in planner evaluation centers on verifying plan feasibility and safety before execution, and selecting the best plan through both in-generation and post-generation strategies. Effective reflection combines automated validation, iterative refinement, and structured multi-path reasoning to enable adaptive and reliable decision-making progress.

\section{How to evaluate? The evaluation methodologies and data}
Evaluating LLM-based agent in multi-turn scenario involves a variety of methodologies, each with its own strengths and limitations. These systems are complex, requiring evaluation methods that can capture their dynamic and interactive nature. \autoref{tab:summary_table_1} and \autoref{tab:summary_table_2} summarize recent related benchmarks and datasets.

\subsection{Evaluation Data}
High-quality evaluation data serves as the prerequisite for robust benchmarking, necessitating a dual focus on constructing realistic conversational scenarios and establishing precise labeling standards to accurately gauge agent capabilities.
Here, we focus on the methodologies for generating and annotating conversation data used to evaluate the LLM-based agents. The evaluation taxonomy categorizes the data into two primary dimensions: \textbf{Conversation Data Generation} and \textbf{Conversation Data Annotation}. In the former, processes include the generation of next turn responses to simulate natural multi-turn interactions, the creation of tool-use data reflecting how agents utilize tools and APIs during conversations, the production of query rewritten data to evaluate the adaptation or reframing of user inputs, and the generation of fact check data to test the agent’s ability to verify and present factual information accurately. In the latter, annotation processes are applied to ensure the generated data can be effectively used for evaluation; these involve annotating the expected next turn response, capturing how tools and function calls are employed by the agent, marking rewritten queries and retrieved items to assess alignment with user intent and data relevance, and verifying factual accuracy through fact checks, as briefly represented in \autoref{fig:evaluation_data_taxonomy}.

\input{evaluation_data_taxonomy}

\subsubsection{Conversation Data Generation}
Automated generation of multi-turn dialogue data reduces manual annotation while ensuring metrics such as turn-level coherence and semantic relevance \cite{jiang-etal-2022-im2}. Key challenges include maintaining contextual consistency and adapting to dynamic intent shifts \cite{chen-etal-2024-diahalu}. Approaches include \textbf{Next Turn Response Data} methods that generate context-aware replies using dialogue state representations via ISU \cite{golany2024efficientdatagenerationsourcegrounded}, hierarchical attention in HMAN \cite{9020160}, chain-of-thought reasoning in CoT Distillation \cite{chae-etal-2023-dialogue}, and adversarial training in hredGAN \cite{olabiyi2019multiturndialogueresponsegeneration}; coherence is further enhanced with Last Utterance-Context Attention \cite{9277775}. Debate frameworks such as DEBATE \cite{kim2024debatedevilsadvocatebasedassessment} along with multi-agent systems \cite{cite-key, abercrombie-batista-navarro-2020-parlvote, 10.1093/oso/9780198849063.003.0005} and judge-based models \cite{moniri2024evaluatingperformancelargelanguage} further refine responses. In addition to standard conversation data generation, advanced techniques for tool-use, query rewriting, and fact checking have emerged as areas of particular interest. Some of representative studies are as follows:
\begin{itemize}
    \item \textbf{Tool-Oriented Dialogue Generation:}  Automated pipelines, such as ToolDial \cite{anonymous2024tooldial}, establish dialogue states and actions directly from API documentation. Additionally, methods generating diverse test scenarios using LLMs and intermediate graphs are highlighted in \citet{arcadinho2024automatedtestgenerationevaluate}, while manually annotated datasets such as DailyDialog \cite{li2017dailydialogmanuallylabelledmultiturn} and Topical Chat \cite{gopalakrishnan2019topical} provide critical evaluation benchmarks. For further insights into tool-based data generation, see \href{https://example.com/tool-based-dialogue}{Tool-Based Dialogue Innovations}.

    \item \textbf{Query Rewritten Techniques in Dialogue Generation:}  Scalable \textbf{Query Rewritten Data} is produced by refining queries through selective contextual integration \cite{ALTHANI2023100025} and self-supervised learning \cite{liu2021conversationalqueryrewritingselfsupervised}. 
  
    \item \textbf{Factual Verification Methods for Dialogue Generation:} Concurrently, \textbf{Fact Check Data} employs veracity prediction with explanation generation \cite{atanasova-etal-2020-generating-fact}, integrating LLM pre-trained knowledge with dynamic evidence retrieval \cite{tian2024webretrievalagentsevidencebased}, automated question generation pipelines \cite{Setty_2024}, and corrector modules to ensure factual consistency \cite{chaudhury-etal-2022-x}. To explore fact-checking methods in dialogue systems, please visit \href{https://example.com/fact-check-ai}{Fact Checking in AI}.
 
\end{itemize}

\subsubsection{Conversation Data Annotation}
Data annotation is crucial for evaluating multi-turn dialogues by labeling user queries, system responses, and conversation flow to create benchmarks that reflect human judgment. Annotation includes \emph{Next Turn Response As Annotation}: Benchmarks such as HUMOD \cite{Merdivan2020HumanAD} (28,500 dialogues) and MMDU \cite{liu2024mmdumultiturnmultiimagedialog} (27 dialogue turns) assess context preservation. \citet{ghazarian-etal-2022-wrong} emphasize measuring context retention, while ConvBench \cite{liu2024convbenchmultiturnconversationevaluation} and BotChat \cite{duan2023botchatevaluatingllmscapabilities} compare LLM outputs against human standards. The dataset in \cite{deng2024multiturninstructionfollowingconversational} transforms single-turn prompts into multi-turn exchanges .\emph{Tool-use and Function Calls in LLMs As annotation}:
ToolBench \cite{xu2023toolmanipulationcapabilityopensource} tests API call generation, and \citet{patil2023gorillalargelanguagemodel} use AST matching to assess hallucination errors. HuggingGPT \cite{shen2023hugginggptsolvingaitasks} orchestrates model interactions, while \citet{liu-etal-2024-evaluation-mechanism} and \citet{li-etal-2023-api} benchmark multi-turn API interactions. MTU-Bench \cite{wang2024mtubenchmultigranularitytoolusebenchmark} evaluates tool usage over multiple turns. \emph{Query Rewritten and Retrieved Items as Annotation}:
\citet{su-etal-2019-improving} collect annotations where human annotators rewrite utterances to clarify intent. In a similar vein, \citet{mo2024aligningqueryrepresentationrewritten} leverage rewritten queries with relevance judgments, while \citet{Aliannejadi_2020} annotate past utterances to identify useful items for interpreting current queries. Lastly \emph{Fact-Checking Annotation in Dialogue Evaluation}
Fact-checking annotations verify accuracy using document-level claim analysis \cite{sathe-park-2021-automatic}. \citet{gupta-etal-2022-dialfact} introduce the DialFact dataset (22k claims with Wikipedia evidence), while multi-stage schemes \cite{wang2024factcheckbenchfinegrainedevaluationbenchmark} enable turn-by-turn evaluation. \citet{zhao-etal-2024-matters} present the BELIEF benchmark, and BERT-based models trained on WikiFactCheck-English \cite{sathe-etal-2020-automated} automate verification.

\input{evaluation_metrics_taxonomy}

\subsection{Evaluation Metrics}
\begin{figure}[h]
    \centering
    \includegraphics[width=0.95\textwidth]{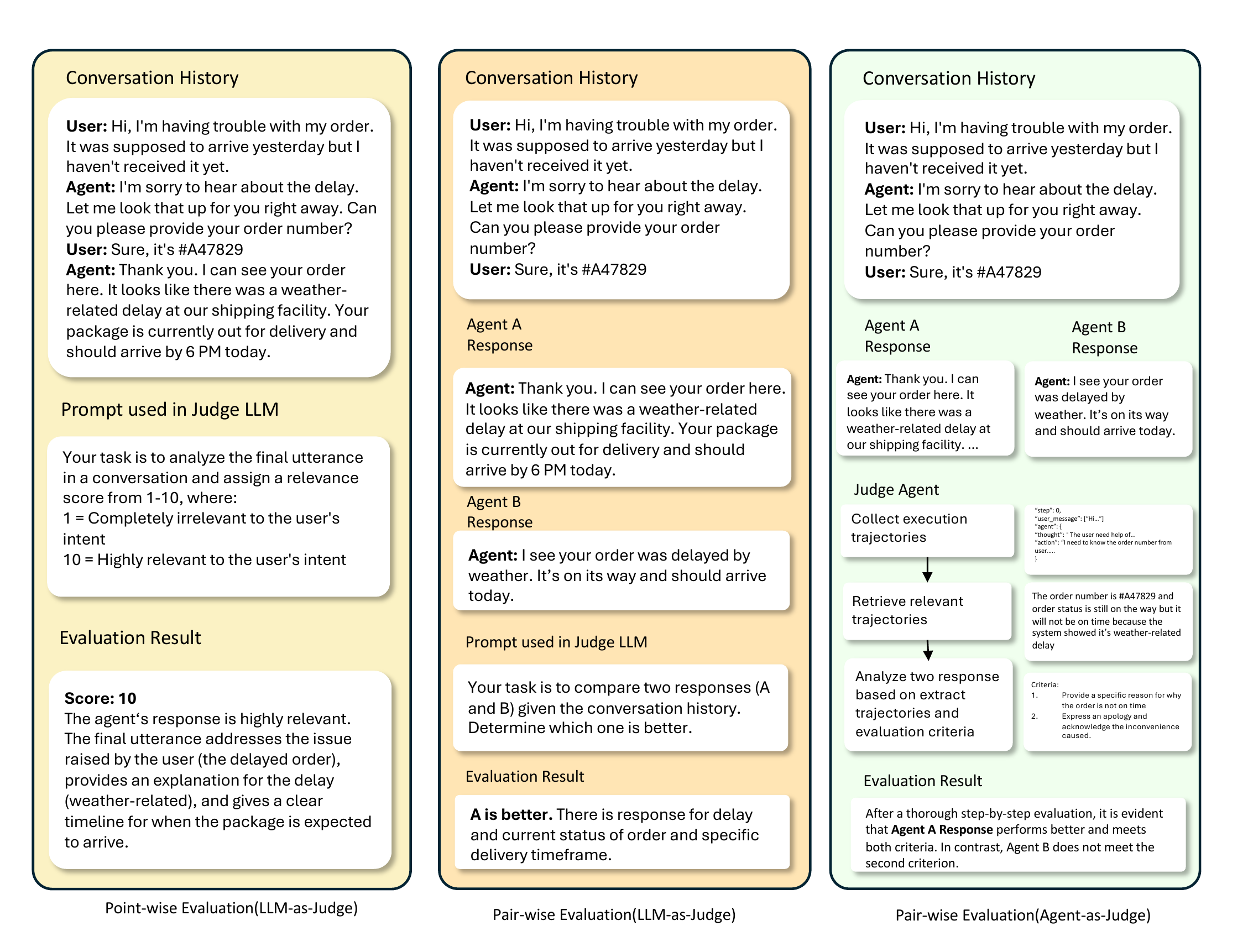}
    \caption{An Example of Point-wise and Pair-wise Evaluation}
    \label{fig:sidebyside_example}\vspace{-1mm}
\end{figure}

Robust evaluation metrics serve as the critical yardstick for quantifying agent intelligence and interaction quality, necessitating a framework that encompasses both rigorous annotation-based benchmarks and scalable annotation-free methodologies.
This section reviews evaluation metrics and methods for assessing multi-turn conversational agents, as briefly illustrated in \autoref{fig:evaluation_metrics} and \autoref{fig:sidebyside_example} provides a concrete example of how these metrics are applied in practice. The evaluation approaches are primarily divided into two categories: \emph{Annotation-based Evaluation}, which relies on pre-labeled data to benchmark responses, and \emph{Annotation-free Evaluation}, which automates scoring without the need for manual labels. Annotation-based methods provide detailed, human-curated insights into dialogue quality but are resource-intensive and can be limited by subjective biases in annotation. In contrast, annotation-free metrics leverage self-supervised learning and real-time adaptation to offer scalable evaluations across diverse dialogue contexts, although they may initially require robust training data for optimal performance. This shift towards automation in evaluation not only reduces costs but also enhances scalability, making it feasible to assess conversational agents in real-world, dynamic environments with greater efficiency and adaptability. A comparison between two method is at Table \ref{tab:annotation-vs-llm-judge}.

\input{evaluation_method_comparison}

\subsubsection{Annotation-based Evaluation}
Annotation-based evaluation relies on expert-curated data as benchmarks for assessing an agent's performance, as demonstrated in recent studies \cite{gusev2024pingpongbenchmarkroleplayinglanguage}. Broadly, annotations serve two key functions:

\begin{itemize}
  \item \textbf{Annotation as a Reference for Quality Evaluation:}  
  Annotations are fundamental for evaluating conversational models. For example, \citet{gusev2024pingpongbenchmarkroleplayinglanguage} introduce a role-playing framework that aligns automated metrics with human judgment, while expert-rated multi-modal annotations \cite{gritta2024humanrankevalautomaticevaluationlms, luo2022alichilargescalemultimodaldataset} enhance evaluation quality. GupShup addresses code-switched conversations \cite{Mehnaz2021GupShupSO}, and methods like Dialog Quality Annotation (DQA) \cite{komma2023accurategeneralizableevaluationmetrics} and benchmarks such as HUMOD \cite{app10030762} underscore the importance of high-quality annotations. Datasets like MultiWOZ \cite{budzianowski-etal-2018-multiwoz} and various persona-based datasets \cite{zhang2018personalizingdialogueagentsi} further ensure model consistency. Traditional metrics such as \emph{BLEU} \cite{10.3115/1073083.1073135, post2018clarityreportingbleuscores}, \emph{ROUGE} \cite{lin-2004-rouge}, and \emph{METEOR} \cite{10.5555/1626355.1626389} have notable limitations. To overcome these, advanced methods leverage semantic and contextual insights using \emph{BERTScore} \cite{zhang2020bertscoreevaluatingtextgeneration, kamal-eddine-etal-2022-frugalscore}, \emph{DialogRPT} \cite{gao2020dialogueresponserankingtraining}, \emph{USL-H} \cite{phy2020deconstructreconstructconfigurableevaluation}, and \emph{Cosine Similarity} \cite{banerjee2023benchmarkingllmpoweredchatbots}. However, one significant limitation of annotation-based methods is the high cost and labor-intensive process of data annotation. This often leads to inconsistencies due to human biases or disagreements in annotation. Furthermore, annotations may not fully capture all dimensions of conversational quality, such as user satisfaction or model creativity, which remain subjective.

  \item \textbf{Annotation as the Exact Result for Matching:}  
  In multi-turn scenarios, exact matching of annotations is essential for high-precision tasks such as \emph{intent recognition} \cite{yang2021ubarfullyendtoendtaskoriented} and \emph{entity extraction} \cite{wu2024benchmarkinglargelanguagemodels}. This strict alignment is vital for reliable conversational AI, with evaluation methods focusing on \emph{context attention distribution} (e.g., using the DAS ratio \cite{xing2022evaluatingimprovingcontextattention}) and \emph{entity verification} (e.g., with Inform \cite{yang2021ubarfullyendtoendtaskoriented}). Comprehensive evaluations also benefit from granular, turn-level annotations provided by resources such as MultiWOZ \cite{budzianowski2020multiwozlargescalemultidomain} and Taskmaster-1 \cite{byrne-etal-2019-taskmaster} \cite{dong2024bamboocomprehensivebenchmarkevaluating, wu2024benchmarkinglargelanguagemodels}. Moreover, as demonstrated in the GAIA benchmark \cite{DBLP:conf/iclr/MialonF0LS24}, anchoring each question to a singular, factual ground truth facilitates an unambiguous exact-match evaluation. Nevertheless, matching annotations precisely in complex dialogues is challenging and may fail to account for diverse conversational contexts, leading to an underestimation of model performance in real-world interactions. Additionally, this strict matching approach limits flexibility, as some nuances in human responses may not match predefined annotation structures.
\end{itemize}

\subsubsection{Annotation-free Evaluation}
Annotation-free evaluation methods, which rely on intrinsic metrics rather than extensive human-annotated datasets, are increasingly critical. While human evaluation remains the gold standard, its high cost and variable agreement underscore the need for more consistent, automated approaches \cite{smith-etal-2022-human, zhang2024comprehensiveanalysiseffectivenesslarge, liu-etal-2022-reference, chiang-lee-2023-closer, chiang-lee-2023-large}.

\begin{itemize}
  \item \textbf{Point-wise Evaluation:}  This approach assigns scores to individual responses based on predefined criteria, making it ideal for large-scale automated assessments \cite{fan-etal-2024-sedareval}. It evaluates factors such as relevance, clarity, and topicality \cite{ferron-etal-2023-meep, lin-chen-2023-llm}. Several techniques have been proposed, including standard metrics \cite{10.1007/978-3-662-49390-8_35}, semi-automated methods \cite{knowledge2010004}, hierarchical residual matching networks \cite{Zhang_2021}, and even using GPT-4 as a judge \cite{li2024fbbenchfinegrainedmultitaskbenchmark}. BERT models have been applied to assess naturalness, fluency, and informativeness \cite{a-2022-automating, shi-etal-2023-rade}. Additionally, Conditional Pointwise Mutual Information (C-PMI) offers a reference-free metric \cite{ren2023cpmiconditionalpointwisemutual}, while a Transformer-based thread encoder supports dialogue extraction and scoring \cite{jia2023multiturnresponseselectionusing}.

  \item \textbf{Pair-wise or List-wise Comparison:}  Instead of scoring responses individually, this method compares multiple responses simultaneously to directly assess accuracy, coherence, and helpfulness \cite{kim-etal-2024-debate, ruiz-dolz-etal-2023-automatic, liu-etal-2024-empirical, DBLP:conf/nips/BaiY0LHWYZXLZLH23}. A head-to-head win-loss-tie framework, based on the Bradley-Terry model \cite{10.1214/aos/1079120141} and TrueSkill ranking \cite{NIPS2006_f44ee263}, facilitates such evaluations \cite{lee2020evaluationprotocolgenerativeconversational}. Methods like PairEval and ACUTE-Eval further refine these comparisons, while semi-automated approaches combine rule-based, retrieval-based, and generative strategies \cite{knowledge2010004}. Additional relevance measures, such as METEOR \cite{banerjee-lavie-2005-meteor} and TF-IDF \cite{Robertson2004UnderstandingID}, have also been employed. Automated frameworks like ChatEval leverage multi-agent debates to provide numerical scores and detailed reasoning \cite{chan2023chatevalbetterllmbasedevaluators}. Recent studies have introduced training-free metrics that enhance semantic matching and align closely with human judgments \cite{10.1007/978-3-662-49390-8_35, Zhang_2021, jia2023multiturnresponseselectionusing, ren2023cpmiconditionalpointwisemutual, knowledge2010004}.
  While these methods effectively evaluate static outputs, they face limitations when assessing agentic systems that employ dynamic reasoning and tool usage. The authors argue that assessing not only final outputs but also the fulfilment of intermediate milestones and dependencies for agent-involved system could yield judgments that align more closely with human consensus when process structure matters. Thus they proposed "agent-as-judge" method for pair-wise comparison \cite{zhuge2024agentasajudgeevaluateagentsagents}. An agent-as-judge evaluation requires two critical capabilities: capture of intermediate process information during agent execution, and criteria-aware retrieval of the most relevant information from this stored process data to serve as judge input. However, this work represents only the beginning of a broader evaluation paradigm, as multi-turn scenarios encompass numerous domains beyond code generation where agent-as-judge could prove valuable, with promising directions including standardizing storage formats for process information and integrating RAG techniques to enhance information retrieval for more systematic and effective evaluation.

\end{itemize}

\subsection{Benchmark Datasets and Resources}
We present the existing resources on the benchmark datasets and evaluation tools in two comprehensive tables. Table~\ref{tab:summary_table_1} details benchmarks focused on task multi-tasking capabilities, interaction patterns, and temporal dimensions, along with safety and tool-use aspects, thereby outlining how varied methodologies address dialogue coherence and context maintenance. Meanwhile, Table~\ref{tab:summary_table_2} presents benchmarks that evaluate the reliability of tool-use, next-turn response quality, query rewriting, conversation memory, complete interaction, memory editing, task representation, context modeling, task decomposition, and plan selection. Together, these tables provide a holistic overview of current evaluation practices, highlighting both the diverse approaches and the critical metrics employed to assess the performance of LLM-based conversational systems.

\input{combined-tables}

\section{Summary, Challenges, and Future Works}
In this section, we first revisit the co-evolution of conversational systems and the evaluation techniques, while pinpointing the limitations and challenges of current systems and outlining promising future research directions.

\subsection{Trends and Status-quo}
\begin{figure}[h]
    \centering
    \includegraphics[width=0.95\textwidth]{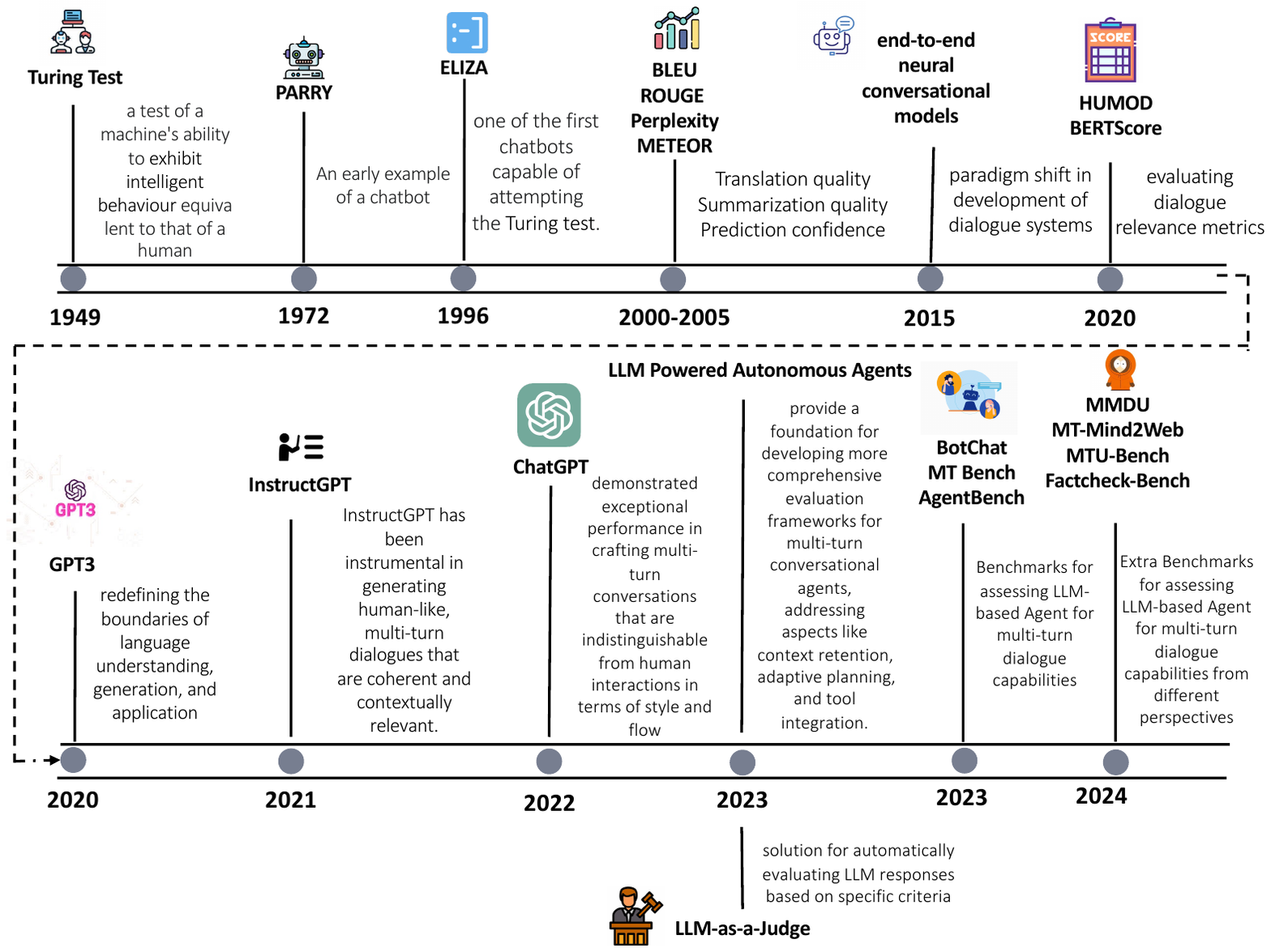}
    \caption{A Roadmap of Evaluation of LLM-based Agents for Multi-turn Conversation. The blue line represents the development of metrics and benchmark for Multi-turn Conversation. The orange line represents the development of Multi-turn Conversation}
    \label{fig:roadmap}\vspace{-3mm}
\end{figure}

\autoref{fig:roadmap} is a concise roadmap highlighting how the evaluation of multi-turn conversational agents has evolved from early rule-based systems to the sophisticated LLM-based agents of today. Three major phases have been experienced by the research community.

\begin{itemize}
    \item \textbf{Early Developments:} Early multi-turn conversation systems were primarily rule-based, with iconic examples including Weizenbaum’s ELIZA (1966) \cite{10.1145/365153.365168} and PARRY (1972) \cite{10.5555/212154.212175}. These systems relied on scripted rules and pattern matching rather than large-scale, data-driven methods. Evaluation methods during this era focused on simple metrics such as user satisfaction surveys and Turing Test–style qualitative assessments, reflecting the constrained nature of the interactions and the lack of robust benchmarks \cite{10.5555/216408.216410}.

    \item \textbf{Transition to Neural Models}: With the advent of deep neural networks, rule-based approaches began to be supplanted by end-to-end neural conversational models. Influential studies such as \citet{sordoni-etal-2015-neural} and \citet{vinyals2015neuralconversationalmodel} introduced models that improved fluency and context understanding. New evaluation metrics like BLEU \cite{10.3115/1073083.1073135} and perplexity \cite{10.1121/1.2016299} emerged to measure generation quality over multiple turns. Researchers started recognizing the limitations of purely reference-based metrics and began incorporating more contextual and diversity-oriented frameworks (e.g., human preference ratings, embedding-based scores).

    \item \textbf{The Rise of LLMs and Agents}: The advent of transformer architectures \cite{10.5555/3295222.3295349} paved the way for large language models, culminating in GPT-style models \cite{Radford2018ImprovingLU} that drastically improved multi-turn conversational capabilities. Key milestones include GPT-2 \cite{Radford2019LanguageMA}, GPT-3 \cite{brown2020languagemodelsfewshotlearners}, followed by instruction-tuned and dialogue-focused systems like InstructGPT \cite{ouyang2022traininglanguagemodelsfollow} and ChatGPT \cite{OpenAI2023ChatGPT}. These models demanded sophisticated, human-in-the-loop evaluation strategies to capture higher-level coherence, factual correctness, and response relevance. Advanced techniques such as CoT prompting \cite{wang2023planandsolvepromptingimprovingzeroshot} refined the reasoning process in multi-turn dialogues, while frameworks like ReAct \cite{yao2022react} and LangChain \cite{Chase_LangChain_2022} enabled improved factuality and task completion through external tool integration.

\end{itemize}
From the early rule-based systems to modern LLM-based agents, the evolution of conversational systems has largely been driven by improvements in fluency, contextual handling, and adaptive response generation. With each developmental phase, evaluation metrics have advanced from simple qualitative assessments to more complex, multi-faceted benchmarks, metrics, and methods aimed at measuring dialogue coherence and context retention. Hereby, we summarize the recent research focus on the evaluation techniques as follows.
\begin{itemize}
    \item \textbf{Holistic Evaluation Benchmarks and Datasets}: As LLMs advanced, the field shifted toward more comprehensive benchmarks that incorporate human judgments, advanced automatic metrics such as BERTScore \cite{zhang2020bertscoreevaluatingtextgeneration, kamal-eddine-etal-2022-frugalscore}, and task-specific evaluations to reflect the increasingly complex, context-rich nature of multi-turn interactions \cite{deng2024multiturninstructionfollowingconversational, arcadinho2024automatedtestgenerationevaluate}. The complexity of multi-turn dialogues led to the creation of specialized datasets and benchmarks that focus on realism, domain knowledge, and long-horizon reasoning: \emph{(1) Dialogue Modeling}: HUMOD \cite{Merdivan2020HumanAD} for human-like dialogue modeling; BotChat \cite{duan2023botchatevaluatingllmscapabilities} for open-ended chat scenarios; AgentBench \cite{liu2023agentbenchevaluatingllmsagents} for agent-centric tasks. \emph{(2) Tool and Action Integration}: API-Bank \cite{li-etal-2023-api} and MMDU \cite{liu2024mmdumultiturnmultiimagedialog} assess real-world tool usage, while MT-Mind2Web \cite{deng2024multiturninstructionfollowingconversational}, MTU-Bench \cite{wang2024mtubenchmultigranularitytoolusebenchmark}, and Factcheck-Bench \cite{wang2024factcheckbenchfinegrainedevaluationbenchmark} provide structured evaluations of multi-turn reasoning, user intent, and factual veracity.

    \item \textbf{LLM/Prompt/Agent-based and Self-Judging Evaluation Methods}: The rise of LLM-based models has accelerated the adoption of annotation-free methods, such as point-wise scoring and side-by-side comparison, reducing dependence on manual evaluation. In some cases, LLMs are employed as evaluators—providing direct numeric scores or explaining their ranking decisions—to enable rapid, iterative testing of multi-turn conversational performance \cite{fan-etal-2024-sedareval, smith-etal-2022-human, zhang2024comprehensiveanalysiseffectivenesslarge, jones-etal-2024-multi}. Learned reward models have emerged as a crucial component in the evaluation and improvement of conversational AI systems, the reward model’s score serves as an approximate “quality meter” for dialogue responses, consolidating many evaluation factors into one number \cite{choi-etal-2024-combining, wang-etal-2024-interpretable, son2024llmasajudgerewardmodel}. While not perfect, these learned models have proven to correlate well with aggregate human judgments in many settings \cite{NEURIPS2020_1f89885d}.

This mapping outlines the historical progression from early rule-based systems through the neural model transition to modern LLM-based agents, and it emphasizes the evolving challenges in evaluation—from simple qualitative metrics to sophisticated, multi-dimensional benchmarks that address the richness of multi-turn interactions.

\end{itemize}

\subsection{Challenges and Future Work}
Despite significant advancements, current evaluation strategies still struggle to fully capture the understanding of multi-turn dialogues, adapting the dynamic interplay between turns rather than assessing them in isolation. Consequently, advances in external tool integration, memory retention, real-time self-assessment, scalability, and privacy preservation are essential to meet the comprehensive requirements of modern dialogue systems. We discuss the limitations of existing studies and propose avenues for future research.

\begin{itemize}
    \item \textbf{Unified and Adaptive Evaluation Frameworks.} Current evaluation methods tend to assess conversation turns in isolation rather than holistically, which limits the ability to capture the dynamic interplay among successive turns. Early systems, relying on simple evaluations such as user satisfaction surveys and Turing Test–style assessments \cite{10.5555/216408.216410}, illustrate this limitation. As dialogue systems have evolved—with neural conversational models \cite{sordoni-etal-2015-neural,vinyals2015neuralconversationalmodel} and later LLM-based approaches \cite{Radford2018ImprovingLU,brown2020languagemodelsfewshotlearners}—there is an increasing need for frameworks that integrate both turn-level and overall conversation assessments. Future research should focus on developing adaptive metrics that can dynamically adjust to variations in context, thereby providing a comprehensive picture of agent performance over extended, multi-turn exchanges.

    \item \textbf{Memory and Context Retention.} Many current benchmarks fail to differentiate between short-term recall and long-term context integration, resulting in issues such as context leakage or drift over prolonged interactions. The early conversational systems often employed rudimentary context tracking techniques, whereas modern neural conversational models still face challenges in maintaining continuity over multiple turns \cite{10.3115/1073083.1073135,10.1121/1.2016299}. Future work should target the creation of specialized benchmarks that accurately measure both temporary and persistent memory retention. This would ensure that agents are capable of leveraging conversational history effectively to maintain coherence and contextual awareness throughout extended dialogues.

    \item \textbf{Test-Time Evaluation for Self-Assessment.} Currently, LLM-based agents generate responses without the capacity to evaluate the quality of their outputs in real time. This lack of inline self-assessment can result in suboptimal responses and missed opportunities for immediate correction, undermining the coherence of extended dialogues. Recent advancements in reasoning techniques such as Chain-of-Thought have opened up new possibilities for iterative evaluation during inference \cite{wang2023planandsolvepromptingimprovingzeroshot}. Future work should focus on integrating test-time evaluation strategies that allow agents to continuously gauge parameters such as coherence, factual accuracy, and context alignment as part of their response generation process. Incorporating real-time feedback loops will support adaptive mechanisms that dynamically correct errors and enhance overall dialogue quality.

    \item \textbf{Dynamic Self-Correction and Error Propagation.} Errors occurring in the initial stages of a conversation can propagate and compound over subsequent turns, leading to incoherent or hallucinated responses. The limitations of early metrics have been highlighted by the evolution from rule-based to neural models \cite{10.5555/216408.216410,vinyals2015neuralconversationalmodel}. Although techniques like Chain-of-Thought prompting have begun refining the reasoning process \cite{wang2023planandsolvepromptingimprovingzeroshot}, there remains a significant gap in real-time error detection and correction. Future research must embed dynamic self-correction mechanisms—such as multi-reasoning trees and iterative feedback loops—that empower agents to continually monitor, reassess, and rectify errors during ongoing interactions.

    \item \textbf{Tool-Use and Action Planning in Extended Interactions.} Modern LLM-based agents now integrate external tools and APIs to accomplish tasks; however, existing evaluations rarely capture how these capabilities evolve throughout a dialogue. Frameworks like ReAct \cite{yao2022react} and platforms such as LangChain \cite{Chase_LangChain_2022} have begun enabling agents to utilize structured knowledge and external resources. Nonetheless, it remains challenging to assess how well an agent adapts its action planning in response to dynamic inputs and shifting task contexts. Future evaluation designs should simulate multi-turn tasks that emphasize cumulative tool-use and adaptive planning, providing systematic insights into how sequential API calls and planning adjustments impact overall task outcomes.

    \item \textbf{Scalability and Real-World Applicability.} The high computational costs and heavy reliance on manual annotations have made it difficult to scale current evaluation methods to real-world scenarios. While early approaches could rely on simplified metrics, modern LLM-based agents require more nuanced and multi-dimensional evaluation strategies to mirror production-level interactions \cite{fan-etal-2024-sedareval,smith-etal-2022-human}. Future directions should focus on automated, annotation-free evaluation pipelines that utilize self-supervised techniques and real-time adaptive metrics.By minimizing reliance on manual annotations, automated systems can leverage large-scale interaction data to derive meaningful metrics. A hierarchical approach, integrating fast initial filters with deeper evaluations, supports efficient processing of extensive dialogues. Integrating evaluation into simulation environments allows for rigorous testing under diverse conditions.

    \item \textbf{Privacy Preservation in Conversational Evaluation.} In multi-turn dialogue systems, preserving user privacy during evaluation is of paramount importance, especially when private conversations between users and agents are involved. Traditional evaluation methods rarely address the risk of exposing sensitive data, a challenge exacerbated by the increased data handling required for complex, multi-turn interactions. Future research must explore methods to evaluate conversation quality in a privacy-preserving manner by incorporating techniques such as Trusted Execution Environment~\cite{jauernig2020trusted,costan2016intel} and federated learning. These approaches should allow rigorous performance assessments while ensuring that individual conversation details remain anonymized and secure, thereby building trust and safeguarding user confidentiality in real-world applications.

    \item \textbf{Philosophical and Ethical Dimensions of Multi-Turn Conversational AI.} The apparent agency of LLM-based agents in multi-turn dialogue forces us to confront the "as-if" attribution of intentionality: although these systems simulate human-like understanding, their lack of genuine consciousness raises fundamental questions about the moral status we ascribe to their outputs and introduces "responsibility gaps" where traditional accountability mechanisms fail~\cite{gumusel2024userprivacyharmsrisks, alberts2024agenticconversationalaichange}. True ethical deployment demands transparency about an agent's non-human nature and probabilistic reasoning, together with robust consent mechanisms that inform users how their data are used and how the system may err, thereby preserving autonomy and preventing deceptive anthropomorphism. Future research must develop frameworks for maintaining human-centered care relationships while leveraging AI capabilities, ensuring that technological advancement enhances rather than replaces the moral foundations of professional practice in sensitive domains.

    \item \textbf{Philosophical and Ethical Dimensions of Multi-Turn Conversational AI.} The perceived agency of LLM-based agents in multi-turn dialogue compels us to confront the “as-if” attribution of intentionality. Although these systems simulate understanding, their lack of consciousness creates responsibility gaps where traditional accountability falters~\cite{gumusel2024userprivacyharmsrisks, alberts2024agenticconversationalaichange}. Ethical evaluation must ensure transparency about an agent’s non-human status, probabilistic reasoning, and internal chain-of-thought. Real-time inspection of reasoning paths against moral and safety standards enables early termination of any turn that violates ethical guidelines. Evaluation frameworks should incorporate metrics to audit thought-process transparency, identify opaque reasoning steps, and assess the effectiveness of moral filters, ensuring AI augments rather than supplants the moral foundations of sensitive practices.

\end{itemize}

\section{Conclusion}
This paper provides a comprehensive overview of evaluation methods for LLM-based agents in multi-turn conversations. By systematically examining the evaluation goals—such as dialogue coherence, effective tool use, and memory retention—and the diverse methodologies, including annotation-based assessments and automated metrics, the study offers a detailed synthesis of the progression from rule-based systems to transformer-based agents. This work lays a solid foundation for future explorations in the domain of conversational AI by addressing both theoretical and practical aspects. The core contribution of this paper lies in developing a structured taxonomy that clarifies critical dimensions of multi-turn conversational performance. This framework not only defines what elements should be evaluated but also establishes the methodological steps necessary to assess these components reliably. By synthesizing insights from nearly 200 scholarly sources, the study bridges the gap between existing evaluation methods and the emerging complexities inherent in modern LLM-based agents.

Additionally, the paper critically addresses current challenges and limitations within the field, such as inadequate benchmarks for long-term memory retention and the need for scalable, annotation-free evaluation pipelines. These contributions highlight the necessity for advanced methodologies that can accurately reflect real-world conversational dynamics and ensure the reliable performance of multi-turn agents. The discussion of future research directions emphasizes the importance of developing automated evaluation techniques and adopting tools that enhance the interpretability and reliability of assessment processes. In summary, this paper not only sets forth a detailed taxonomy and systematic analysis of various evaluation methods but also serves as a catalyst for future research. The study's thorough review of literature, combined with its critical insights into existing challenges, paves the way for the development of more robust evaluation tools, ultimately advancing the field of conversational AI.

\bibliographystyle{ACM-Reference-Format}
\bibliography{sample-base}

\clearpage
\end{document}

%% file: taxonomy.tex
\tikzstyle{my-box}=[
    rectangle,
    draw=hidden-draw,
    rounded corners,
    text opacity=1,
    minimum height=1.5em,
    minimum width=5em,
    inner sep=2pt,
    align=center,
    fill opacity=.5,
    line width=0.8pt,
]
\tikzstyle{leaf}=[my-box, minimum height=1.5em,
    fill=hidden-pink!80, text=black, align=left,font=\normalsize,
    inner xsep=2pt,
    inner ysep=4pt,
    line width=0.8pt,
]
\begin{figure*}[t!]
    \centering
    \resizebox{0.98\textwidth}{!}{
        \begin{forest}
            forked edges,
            for tree={
                grow=east,
                reversed=true,
                anchor=base west,
                parent anchor=east,
                child anchor=west,
                base=center,
                font=\large,
                rectangle,
                draw=hidden-draw,
                rounded corners,
                align=left,
                text centered,
                minimum width=14em,
                edge+={darkgray, line width=1pt},
                s sep=3pt,
                inner xsep=2pt,
                inner ysep=3pt,
                line width=0.8pt,
                ver/.style={rotate=90, child anchor=north, parent anchor=south, anchor=center},
            },
            where level=1{text width=10em,font=\normalsize,}{},
            where level=2{text width=10em,font=\normalsize,}{},
            where level=3{text width=12em,font=\normalsize,}{},
            where level=4{text width=10em,font=\normalsize,}{},
            [
                A Survey On Evaluating LLM-based Agents for Multi-Turn Conversations, ver
                [
                    What to evaluate?\\The evaluation goals \\and target
                    [
                        Evaluating the\\End-to-end Experience 
                        [
                            \citet{Reimann2023PredictingIQ}{, }\citet{Deriu_2020}{, }\citet{Hendrycks2020MeasuringMM}{, }
                            \\
                            \citet{gritta2024humanrankevalautomaticevaluationlms}{, }\citet{Zheng2023JudgingLW}{, }\citet{kwan2024mtevalmultiturncapabilitiesevaluation}{, }\citet{Bai_2024}{, }
                            \\
                            \citet{xu2021goldfishmemorylongtermopendomain}{, }
                            \citet{jang2023conversation}{, }\citet{maharana2024evaluatinglongtermconversationalmemory}{, }
                            \\
                            \citet{liu2024convbenchmultiturnconversationevaluation}{, }\citet{yu2024cosafeevaluatinglargelanguage}
                            , leaf, text width = 32em
                        ]
                    ]
                    [
                        Evaluating the Action\\Tool{-}use Components 
                        [
                            \citet{openai_chatgpt_plugins}{, }\citet{press2023measuringnarrowingcompositionalitygap}{, }\citet{gao2023palprogramaidedlanguagemodels}{, }
                            \citet{xu2023toolmanipulationcapabilityopensource}{, }
                            \\
                            \citet{shen2023hugginggptsolvingaitasks}{, }\citet{patil2023gorillalargelanguagemodel}{, }
                            \citet{wang2024mtubenchmultigranularitytoolusebenchmark}{, }\citet{wang2024mintevaluatingllmsmultiturn}{, }
                            \\
                            \citet{guo-etal-2024-ctooleval}{, }
                            \citet{zhuang2023toolqadatasetllmquestion}{, }\citet{zhang2024toolbehonestmultilevelhallucinationdiagnostic}
                            , leaf, text width = 32em
                        ]
                    ]
                    [
                        Evaluating the memory 
                        [
                            Memory Spans
                            [
                            \citet{Wang2020DualDM}{, }\citet{Shen2022KWickChatAM}{, }\citet{Cai2022MemoryGW}{, }\citet{leng2024longcontextragperformance}{, }
                            \\
                            \citet{maharana2024evaluatinglongtermconversationalmemory}{, }\citet{zhang2024memsimbayesiansimulatorevaluating}{, }\citet{liu2024llmconversationalagentmemory}{, }
                            \\
                            \citet{castillobolado2024promptsdynamicconversationalbenchmarking}{, }\citet{huang2023memorysandboxtransparentinteractive}
                            , leaf, text width = 32em
                            ]
                        ]
                        [
                            Memory Forms
                            [
                             \citet{zhang2024surveymemorymechanismlarge}{, }\citet{zhang2023historyawarehierarchicaltransformermultisession}{, }\citet{wang2023fccfusingconversationhistory}{, }\\
                            \citet{huang2024doesconversationlengthimpact}{, }\citet{li2023how}{, }\citet{maharana2024evaluatinglongtermconversationalmemory}{, }\\
                            \citet{lei2024s3evalsyntheticscalablesystematic}{, }\citet{pal2023giraffeadventuresexpandingcontext}{, }\citet{tworkowski2023focusedtransformercontrastivetraining}{, }\\
                            \citet{bae2022updatedmemorymanagementlongterm}{, }\citet{Liang2023UnleashingII}{, }\citet{Hoxha2024BlockchainAA}{, }\\
                            \citet{cao2024diaggptllmbasedmultiagentdialogue}{, }\citet{wang2024userbehaviorsimulationlarge}{, }\citet{park2023generativeagentsinteractivesimulacra}{, }\\
                            \citet{ong2024lifelongdialogueagentsrelationaware}{, }\citet{zhong2023memorybankenhancinglargelanguage}{, }\citet{liu2023thinkinmemoryrecallingpostthinkingenable}{, }\\
                            \citet{wang-etal-2024-recmind}{, }\citet{modarressi2024retllmgeneralreadwritememory}{, }\citet{xu2023toolmanipulationcapabilityopensource}{, }\\
                            \citet{qin2023toolllmfacilitatinglargelanguage}{, }\citet{zhang2024personalizedllmresponsegeneration}{, }\citet{li2022largelanguagemodelscontrollable}{, }\\
                            \citet{tan2024democratizinglargelanguagemodels}{, }\citet{fountas2024humanlikeepisodicmemoryinfinite}{, }\citet{mitchell2022memorybasedmodeleditingscale}{, }\\
                            \citet{qiao-etal-2024-comem}{, }\citet{zeng2024famefactualmultitaskmodel}{, }\citet{tack2024onlineadaptationlanguagemodels}{, }\citet{mao2024editingpersonalitylargelanguage}
                            , leaf, text width = 32em
                            ]
                        ]
                    ]
                    [
                        Evaluating the planner
                        [
                            Task Modeling
                            [
                            \citet{2024multi-agent}{, }\citet{singh2024twostepmultiagenttaskplanning}{, }\citet{zhang2024lammapgeneralizablemultiagentlonghorizon}{, }\\
\citet{lyu2022primaplannerreasonerinsidemultitask}{, }\citet{10.1145/3442381.3449902}{, }\citet{xing2022evaluatingimprovingcontextattention}{, }\\
\citet{li2024dialogueactiontokenssteering}{, }\citet{DBLP:conf/kdd/FuZ023}{, }\citet{wang2024metadtofflinemetarlconditional}
, leaf, text width = 32em
                            ]
                        ]
                        [
                            Task Decomposition
                            [
                            \citet{huang2024understandingplanningllmagents}{, }\citet{wang2024tdagmultiagentframeworkbased}{, }\citet{shen2023hugginggptsolvingaitasks}{, }\\
\citet{wang2023planandsolvepromptingimprovingzeroshot}{, }\citet{singh2022progpromptgeneratingsituatedrobot}{, }\citet{srinivasan2023multiagentcollectiveconstructionusing}{, }\\
\citet{chen2023programthoughtspromptingdisentangling}{, }\citet{zhang2024dynamicplanningllmbasedgraphical}{, }\citet{wu2024mldtmultileveldecompositioncomplex}{, }\\
\citet{prasad2024adaptasneededdecompositionplanning}{, }\citet{dagan2023dynamicplanningllm}{, }\citet{kwon2024fastaccuratetaskplanning}, leaf, text width = 32em
                            ]
                        ]
                        [
                            Adaptation and Control
                            [
                            \citet{9864802}{, }\citet{10054388}{, }\citet{hao2023intentdialintentgraphbased}{, }\\
\citet{Huang_2017}{, }\citet{muise2019planninggoalorienteddialoguesystems}, leaf, text width = 32em
                            ]
                        ]
                        [
                            Reflection
                            [
                            \citet{yao2022react}{, }\citet{zhang2024probingmultiturnplanningcapabilities}{, }\citet{10.1145/3287921.3287947}{, }\\
\citet{zheng2024criticcotboostingreasoningabilities}{, }\citet{madaan2023selfrefineiterativerefinementselffeedback}{, }\citet{2024multi-agent}{, }\\
\citet{li2024dialogueactiontokenssteering}{, }\citet{shi2024directmultiturnpreferenceoptimization}{, }\citet{liu2024dynamicllmpoweredagentnetwork}{, }\\
\citet{sun2023adaplanneradaptiveplanningfeedback}{, }\citet{wang2024planningnaturallanguageimproves}{, }\citet{10.1007/978-3-031-48421-6_7}{, }\\
\citet{light2024strategistlearningstrategicskills}{, }\citet{yao2023tree}{, }\citet{Besta_2024}{, }\citet{zhao2023largelanguagemodelscommonsense}{, }\\
\citet{xiao2024llmahumanloop}{, }\citet{he2016deepreinforcementlearningnatural}, leaf, text width = 32em
                            ]
                        ]
                    ]
                ]
                [
                    How to evaluate? \\ Data and Metrics 
                    [
                        Evaluation Data
                        [
                            Conversation Data Generation
                            [
                             \citet{9020160}{, }\citet{olabiyi2019multiturndialogueresponsegeneration}{, }\citet{9277775}{, }\\
\citet{abercrombie-batista-navarro-2020-parlvote}{, }\citet{10.1093/oso/9780198849063.003.0005}{, }\citet{kim2024debatedevilsadvocatebasedassessment}{, }\\
\citet{moniri2024evaluatingperformancelargelanguage}
 \citet{anonymous2024tooldial}{, }\citet{golany2024efficientdatagenerationsourcegrounded}{, }\citet{arcadinho2024automatedtestgenerationevaluate}{, }\\
\citet{li2017dailydialogmanuallylabelledmultiturn}{, }\citet{gopalakrishnan2019topical}\\
 \citet{ALTHANI2023100025}{, }\citet{liu2021conversationalqueryrewritingselfsupervised}\\
 \citet{atanasova-etal-2020-generating-fact}{, }\citet{tian2024webretrievalagentsevidencebased}{, }
\citet{Setty_2024},leaf, text width = 32em
                            ]
                        ]
                        [
                            Conversation Data Annotation
                            [
                             \citet{Merdivan2020HumanAD}{, }\citet{liu2024mmdumultiturnmultiimagedialog}{, }\citet{deng2024multiturninstructionfollowingconversational}{, }\\
\citet{liu2024convbenchmultiturnconversationevaluation}{, }\citet{duan2023botchatevaluatingllmscapabilities}\\
\citet{xu2023toolmanipulationcapabilityopensource}{, }\citet{shen2023hugginggptsolvingaitasks}{, }\citet{wang2024mtubenchmultigranularitytoolusebenchmark}{, }\\
\citet{patil2023gorillalargelanguagemodel}
\\ \citet{su-etal-2019-improving}{, }\citet{mo2024aligningqueryrepresentationrewritten}{, }\\
\citet{Aliannejadi_2020}
 \citet{sathe-park-2021-automatic}{, }\citet{wang2024factcheckbenchfinegrainedevaluationbenchmark}, leaf, text width = 32em
                            ]
                        ]
                    ]
                    [
                        Evaluation Metrics
                        [
                            Annotation-based Evaluation
                            [
                            \citet{10.3115/1073083.1073135}{, }\citet{post2018clarityreportingbleuscores}{, }
\citet{lin-2004-rouge}{, }\citet{10.5555/1626355.1626389}{,} \\
\citet{zhang2020bertscoreevaluatingtextgeneration}{, }\citet{gao2020dialogueresponserankingtraining}{, }
\citet{phy2020deconstructreconstructconfigurableevaluation}{, }\citet{banerjee2023benchmarkingllmpoweredchatbots}\\
\citet{xing2022evaluatingimprovingcontextattention}{, }\citet{10.1145/3545570}{, }\\
\citet{yang2021ubarfullyendtoendtaskoriented}{, }\citet{dong2024bamboocomprehensivebenchmarkevaluating}{, }
\citet{wu2024benchmarkinglargelanguagemodels}{, }\\
\citet{budzianowski2020multiwozlargescalemultidomain}{, }\citet{byrne-etal-2019-taskmaster},leaf, text width = 32em
                            ]
                        ]
                        [
                            Annotation-free Evaluation
                            [
                            \citet{10.1007/978-3-662-49390-8_35}{, }\citet{knowledge2010004}{, }\citet{Zhang_2021}{, }\\
\citet{li2024fbbenchfinegrainedmultitaskbenchmark}{, }\citet{ren2023cpmiconditionalpointwisemutual}{, }
\citet{jia2023multiturnresponseselectionusing}{, }\\
\citet{lee2020evaluationprotocolgenerativeconversational}{, }\citet{park2024pairevalopendomaindialogueevaluation}{, }\\
\citet{li2019acuteevalimproveddialogueevaluation}{, }\citet{knowledge2010004}{, }
\citet{chan2023chatevalbetterllmbasedevaluators},leaf, text width = 32em
                            ]
                        ]
                    ]
                ]
            ]
        \end{forest}
    }
    \caption{Taxonomy of Evaluation Approaches for LLM-Based Multi-Turn Conversational Agents: A Comprehensive Survey of Goals, Methodologies, and Future Directions.}
    \label{fig:taxonomy}\vspace{-5mm}
\end{figure*}

%% file: memory_taxonomy.tex
\tikzstyle{my-box}=[
    rectangle,
    draw=hidden-draw,
    rounded corners,
    text opacity=1,
    minimum height=1.5em,
    minimum width=5em,
    inner sep=2pt,
    align=center,
    fill opacity=.5,
    line width=0.8pt,
]
\tikzstyle{leaf}=[my-box, minimum height=1.5em,
    fill=hidden-pink!80, text=black, align=left, font=\normalsize,
    inner xsep=2pt,
    inner ysep=4pt,
    line width=0.8pt,
]
\begin{figure*}[t!]
    \centering
    \resizebox{0.98\textwidth}{!}{
        \begin{forest}
            forked edges,
            for tree={
                grow=east,
                reversed=true,
                anchor=base west,
                parent anchor=east,
                child anchor=west,
                base=center,
                font=\large,
                rectangle,
                draw=hidden-draw,
                rounded corners,
                align=left,
                text centered,
                minimum width=12em,
                edge+={darkgray, line width=1pt},
                s sep=3pt,
                inner xsep=2pt,
                inner ysep=3pt,
                line width=0.8pt,
                ver/.style={rotate=90, child anchor=north, parent anchor=south, anchor=center},
            },
            where level=1{font=\normalsize, text width=11em}{},
            where level=2{font=\normalsize, text width=10em}{},
            where level=3{font=\normalsize, text width=9em}{},
            where level=4{font=\normalsize, text width=9em}{},
            [
                Memory
                [
                    Memory Spans
                    [
                        Turn Memory
                        [
                            \citet{Wang2020DualDM}{,} \citet{Shen2022KWickChatAM}{,} \citet{Cai2022MemoryGW}, leaf, text width = 22em
                        ]
                    ]
                    [
                        Conversation Memory
                        [
                            \citet{leng2024longcontextragperformance}{,} \citet{chen-etal-2024-socialbench}{,} \\
                            \citet{maharana2024evaluatinglongtermconversationalmemory}{,} \citet{zhang2024memsimbayesiansimulatorevaluating}{,} \\
                            \citet{liu2024llmconversationalagentmemory}, leaf, text width = 18em
                        ]
                    ]
                    [
                        Permanent Memory
                        [
                            \citet{castillobolado2024promptsdynamicconversationalbenchmarking}{,} \citet{huang2023memorysandboxtransparentinteractive}{,}\\
                            \citet{fu-etal-2024-msi}, leaf, text width = 20em
                        ]
                    ]
                ]
                [
                    Memory Forms
                    [
                        Textual Form
                        [
                            Complete Interaction
                            [
                                \citet{zhang2023historyawarehierarchicaltransformermultisession}{,} \citet{wang2023fccfusingconversationhistory}{,} \\
                                \citet{huang2024doesconversationlengthimpact}{,} \citet{li2023how}{,} \\
                                \citet{maharana2024evaluatinglongtermconversationalmemory}{,} \citet{lei2024s3evalsyntheticscalablesystematic}{,} \\
                                \citet{pal2023giraffeadventuresexpandingcontext}{,} \citet{tworkowski2023focusedtransformercontrastivetraining}{,} \\
                                \citet{bae2022updatedmemorymanagementlongterm}, leaf, text width = 18em
                            ]
                        ]
                        [
                            Recent Interaction
                            [
                                \citet{Liang2023UnleashingII}{,} \citet{cao2024diaggptllmbasedmultiagentdialogue}{,} \\
                                \citet{Hoxha2024BlockchainAA}{,}\\
                                \citet{wang2024userbehaviorsimulationlarge}, leaf, text width = 16em
                            ]
                        ]
                        [
                            Retrieved Interaction
                            [
                                \citet{sarch-etal-2023-open}{,} \citet{park2023generativeagentsinteractivesimulacra}{,} \\
                                \citet{ong2024lifelongdialogueagentsrelationaware}{,} \citet{zhong2023memorybankenhancinglargelanguage}{,} \\
                                \citet{johnson2017billionscalesimilaritysearchgpus}{,} \citet{liu2023thinkinmemoryrecallingpostthinkingenable}{,} \\
                                \citet{wang-etal-2024-recmind}{,} \citet{modarressi2024retllmgeneralreadwritememory}, leaf, text width = 18em
                            ]
                        ]
                        [
                            External Interaction
                            [
                                \citet{xu2023toolmanipulationcapabilityopensource}{,} \citet{qin2023toolllmfacilitatinglargelanguage}, leaf, text width = 16em
                            ]
                        ]
                    ]
                    [
                        Parametric Form
                        [
                            Fine-tuned Memory
                            [
                                \citet{zhang2024personalizedllmresponsegeneration}{,} \citet{li2022largelanguagemodelscontrollable}{,} \\
                                \citet{tan2024democratizinglargelanguagemodels}{,} \citet{fountas2024humanlikeepisodicmemoryinfinite}, leaf, text width = 16em
                            ]
                        ]
                        [
                            Memory-editing
                            [
                                \citet{mitchell2022memorybasedmodeleditingscale}{,} \citet{qiao-etal-2024-comem}{,} \\
                                \citet{zeng2024famefactualmultitaskmodel}{,} \citet{tack2024onlineadaptationlanguagemodels}{,} \\
                                \citet{mao2024editingpersonalitylargelanguage}, leaf, text width = 16em
                            ]
                        ]
                    ]
                ]
            ]
        \end{forest}
    }
    \caption{Taxonomy of Evaluation Memory of LLM-Based Agents in Multi-Turn Conversations.}
    \label{fig:memory taxonomy}
    \vspace{-5mm}
\end{figure*}

%% file: planner_taxonomy.tex
\tikzstyle{my-box}=[
    rectangle,
    draw=hidden-draw,
    rounded corners,
    text opacity=1,
    minimum height=1.5em,
    minimum width=5em,
    inner sep=2pt,
    align=center,
    fill opacity=.5,
    line width=0.8pt,
]
\tikzstyle{leaf}=[my-box, minimum height=1.5em,
    fill=hidden-pink!80, text=black, align=left, font=\normalsize,
    inner xsep=2pt,
    inner ysep=4pt,
    line width=0.8pt,
]
\begin{figure*}[t!]
    \centering
    \resizebox{0.98\textwidth}{!}{
        \begin{forest}
            forked edges,
            for tree={
                grow=east,
                reversed=true,
                anchor=base west,
                parent anchor=east,
                child anchor=west,
                base=center,
                font=\large,
                rectangle,
                draw=hidden-draw,
                rounded corners,
                align=left,
                text centered,
                minimum width=12em,
                edge+={darkgray, line width=1pt},
                s sep=3pt,
                inner xsep=2pt,
                inner ysep=3pt,
                line width=0.8pt,
                ver/.style={rotate=90, child anchor=north, parent anchor=south, anchor=center},
            },
            where level=1{font=\normalsize, text width=11em}{},
            where level=2{font=\normalsize, text width=10em}{},
            where level=3{font=\normalsize, text width=9em}{},
            where level=4{font=\normalsize, text width=9em}{},
            [
                Planner
                [
                    Task Modeling
                    [
                         Task Representation
                         [
                            \citet{2024multi-agent}{,} \\
                            \citet{singh2024twostepmultiagenttaskplanning}{,} \\
                            \citet{zhang2024lammapgeneralizablemultiagentlonghorizon}{,} \citet{lyu2022primaplannerreasonerinsidemultitask}, leaf, text width = 16em
                         ]
                    ]
                    [
                        Context Modeling
                        [
                            \citet{feng-etal-2022-represent}{,} \citet{10.1145/3442381.3449902}{,}\\
                            \citet{xiao-etal-2024-flowbench}{,} \citet{li2024dialogueactiontokenssteering}{,} \citet{DBLP:conf/kdd/FuZ023}{,} \\
                            \citet{xing2022evaluatingimprovingcontextattention}{,} \citet{wang2024metadtofflinemetarlconditional}, leaf, text width = 20em
                        ]
                    ]
                ]
                [
                    Task Decomposition
                    [
                        \citet{huang2024understandingplanningllmagents}{,} \citet{wang2024tdagmultiagentframeworkbased}{,} \\
                        \citet{shen2023hugginggptsolvingaitasks}{,} \citet{wang2023planandsolvepromptingimprovingzeroshot}{,} \\
                        \citet{singh2022progpromptgeneratingsituatedrobot}{,} \citet{srinivasan2023multiagentcollectiveconstructionusing}{,} \\
                        \citet{chen2023programthoughtspromptingdisentangling}{,} \citet{zhang2024dynamicplanningllmbasedgraphical}{,} \\
                        \citet{wu2024mldtmultileveldecompositioncomplex}{,} \citet{prasad2024adaptasneededdecompositionplanning}{,} \\
                        \citet{dagan2023dynamicplanningllm}{,} \citet{kwon2024fastaccuratetaskplanning}, leaf, text width = 18em
                    ]
                ]
                [
                    Adaptation and Control
                    [
                      \citet{9864802}{,} \\
                      \citet{10054388}{,} \\
                      \citet{hao2023intentdialintentgraphbased}{,} \citet{Huang_2017}{,} \\
                      \citet{liu-etal-2024-lara}{,} \citet{muise2019planninggoalorienteddialoguesystems}, leaf, text width = 16em
                    ]
                ]
                [
                  Reflection
                  [
                    Plan Verification
                    [
                        \citet{yao2022react}{,} \citet{zhang2024probingmultiturnplanningcapabilities}{,} \citet{10.1145/3287921.3287947}{,} \\
                        \citet{zheng2024criticcotboostingreasoningabilities}{,} \citet{madaan2023selfrefineiterativerefinementselffeedback}{,} \citet{cao-2024-graphreason}, leaf, text width = 22em
                    ]
                  ]
                  [
                    Plan Selection
                    [
                      In-generation Selection, text width = 10em
                      [
                        \citet{cheng-etal-2022-improving}{,} \citet{2024multi-agent}{,} \\
                        \citet{shi2024directmultiturnpreferenceoptimization}{,} \citet{liu2024dynamicllmpoweredagentnetwork}{,}\citet{li2024dialogueactiontokenssteering}{,} \\
                        \citet{sun2023adaplanneradaptiveplanningfeedback}{,} \citet{wang2024planningnaturallanguageimproves}{,} \\
                        \citet{10.1007/978-3-031-48421-6_7}{,} \citet{light2024strategistlearningstrategicskills}, leaf, text width = 20em
                      ]
                    ]
                    [
                      Post-generation Selection, text width = 12em
                      [
                        \citet{yao2023tree}{,} \citet{Besta_2024}{,} \\
                        \citet{kargupta-etal-2024-instruct}{,} \citet{xiao2024llmahumanloop}{,} \\
                        \citet{he2016deepreinforcementlearningnatural}{,} \citet{zhao2023largelanguagemodelscommonsense}{,} \citet{liu-etal-2023-mtgp}, leaf, text width = 20em
                      ]
                    ]
                  ]
                ]
            ]
        \end{forest}
    }
    \caption{Taxonomy of Evaluation Planner of LLM-Based Agents in Multi-Turn Conversations.}
    \label{fig:planner taxonomy}
    \vspace{-5mm}
\end{figure*}

%% file: evaluation_data_taxonomy.tex
\tikzstyle{my-box}=[
    rectangle,
    draw=hidden-draw,
    rounded corners,
    text opacity=1,
    minimum height=1.5em,
    minimum width=5em,
    inner sep=2pt,
    align=center,
    fill opacity=.5,
    line width=0.8pt,
]
\tikzstyle{leaf}=[my-box, minimum height=1.5em,
    fill=hidden-pink!80, text=black, align=left, font=\normalsize,
    inner xsep=2pt,
    inner ysep=4pt,
    line width=0.8pt,
]
\begin{figure*}[t!]
    \centering
    \resizebox{0.98\textwidth}{!}{
        \begin{forest}
            forked edges,
            for tree={
                grow=east,
                reversed=true,
                anchor=base west,
                parent anchor=east,
                child anchor=west,
                base=center,
                font=\large,
                rectangle,
                draw=hidden-draw,
                rounded corners,
                align=left,
                text centered,
                minimum width=12em,
                edge+={darkgray, line width=1pt},
                s sep=3pt,
                inner xsep=2pt,
                inner ysep=3pt,
                line width=0.8pt,
                ver/.style={rotate=90, child anchor=north, parent anchor=south, anchor=center},
            },
            where level=1{font=\normalsize, text width=15em}{},
            where level=2{font=\normalsize, text width=18em}{},
            where level=3{font=\normalsize, text width=9em}{},
            where level=4{font=\normalsize, text width=9em}{},
            [
              Evaluation Data
                [
                    Conversation Data Generation
                    [
                        Generation of Next Turn Response
                        [
                            \citet{9020160}{,} \citet{chae-etal-2023-dialogue}{,} \\
                            \citet{olabiyi2019multiturndialogueresponsegeneration}{,} \\
                            \citet{abercrombie-batista-navarro-2020-parlvote}{,}\\
                            \citet{9277775}{,}  
                            \citet{10.1093/oso/9780198849063.003.0005}{,} \\
                            \citet{kim2024debatedevilsadvocatebasedassessment}{,} \citet{moniri2024evaluatingperformancelargelanguage}, leaf, text width = 16em
                         ]
                    ]
                    [
                        Generation of Tool-Use Data
                        [
                            \citet{anonymous2024tooldial}{,} \citet{golany2024efficientdatagenerationsourcegrounded}{,} \\
                            \citet{arcadinho2024automatedtestgenerationevaluate}{,} \citet{li2017dailydialogmanuallylabelledmultiturn}{,} \\
                            \citet{gopalakrishnan2019topical}, leaf, text width = 16em
                        ]
                    ]
                    [
                        Generation of Query Rewritten Data
                        [
                            \citet{ALTHANI2023100025}{,} \citet{liu2021conversationalqueryrewritingselfsupervised}, leaf, text width = 16em
                        ]
                    ]
                    [
                        Generation of Fact Check Data
                        [
                        \citet{atanasova-etal-2020-generating-fact}{,} \citet{tian2024webretrievalagentsevidencebased}{,} \\
                        \citet{Setty_2024}{,} \citet{chaudhury-etal-2022-x}, leaf, text width = 20em
                      ]
                    ]
                ]
                [
                    Conversation Data Annotation
                    [
                        Next Turn Response As The Annotation
                        [
                        \citet{Merdivan2020HumanAD}{,} \citet{liu2024mmdumultiturnmultiimagedialog}{,} \\
                        \citet{deng2024multiturninstructionfollowingconversational}{,} \citet{ghazarian-etal-2022-wrong}{,} \\
                        \citet{liu2024convbenchmultiturnconversationevaluation}{,} \citet{duan2023botchatevaluatingllmscapabilities}, leaf, text width = 16em
                        ]
                    ]
                    [
                        Tool-Use and Function Calls in LLMs \\As the Annotation
                        [
                        \citet{xu2023toolmanipulationcapabilityopensource}{,} \citet{patil2023gorillalargelanguagemodel}{,} \\
                        \citet{shen2023hugginggptsolvingaitasks}{,} \citet{liu-etal-2024-evaluation-mechanism}{,} \\
                        \citet{li-etal-2023-api}{,} \citet{wang2024mtubenchmultigranularitytoolusebenchmark}, leaf, text width = 16em
                        ]
                    ]
                    [
                        Query Rewritten and Retrieved Items \\As The Annotation
                        [
                        \citet{su-etal-2019-improving}{,} \citet{mo2024aligningqueryrepresentationrewritten}{,} \\
                        \citet{Aliannejadi_2020}, leaf, text width = 16em
                        ]
                    ]
                    [
                        Fact Check Data As The Annotation
                        [
                        \citet{sathe-park-2021-automatic}{,} \citet{gupta-etal-2022-dialfact}{,} \\
                        \citet{wang2024factcheckbenchfinegrainedevaluationbenchmark}{,} \citet{zhao-etal-2024-matters}{,} \\
                        \citet{sathe-etal-2020-automated}, leaf, text width = 16em
                        ]
                    ]
                ]
            ]
        \end{forest}
    }
    \caption{Taxonomy of Evaluation Data for LLM-based Agents in Multi-Turn Conversations.}
    \label{fig:evaluation_data_taxonomy}
    \vspace{-5mm}
\end{figure*}

%% file: evaluation_metrics_taxonomy.tex
\tikzstyle{my-box}=[
    rectangle,
    draw=hidden-draw,
    rounded corners,
    text opacity=1,
    minimum height=1.5em,
    minimum width=5em,
    inner sep=2pt,
    align=center,
    fill opacity=.5,
    line width=0.8pt,
]
\tikzstyle{leaf}=[my-box, minimum height=1.5em,
    fill=hidden-pink!80, text=black, align=left, font=\normalsize,
    inner xsep=2pt,
    inner ysep=4pt,
    line width=0.8pt,
]
\begin{figure*}[t!]
    \centering
    \resizebox{0.98\textwidth}{!}{
        \begin{forest}
            forked edges,
            for tree={
                grow=east,
                reversed=true,
                anchor=base west,
                parent anchor=east,
                child anchor=west,
                base=center,
                font=\large,
                rectangle,
                draw=hidden-draw,
                rounded corners,
                align=left,
                text centered,
                minimum width=12em,
                edge+={darkgray, line width=1pt},
                s sep=3pt,
                inner xsep=2pt,
                inner ysep=3pt,
                line width=0.8pt,
                ver/.style={rotate=90, child anchor=north, parent anchor=south, anchor=center},
            },
            where level=1{font=\normalsize, text width=11em}{},
            where level=2{font=\normalsize, text width=10em}{},
            where level=3{font=\normalsize, text width=9em}{},
            where level=4{font=\normalsize, text width=9em}{},
            [
              Evaluation Metrics
                [
                  Annotation-based Evaluation, text width = 15em
                    [
                         Annotation As A Reference for Quality Evaluation, text width = 22em
                        [
                          Traditional Metrics
                          [
                            \citet{10.3115/1073083.1073135}{,} \citet{post2018clarityreportingbleuscores}{,} \\  
\citet{lin-2004-rouge}{,} \citet{10.5555/1626355.1626389}, leaf, text width = 16em
                          ]
                        ] 
                        [
                          Advanced Metrics
                          [
                            \citet{zhang2020bertscoreevaluatingtextgeneration}{,} \citet{kamal-eddine-etal-2022-frugalscore}{,} \\  
\citet{gao2020dialogueresponserankingtraining}{,} \citet{phy2020deconstructreconstructconfigurableevaluation}{,} \\
\citet{banerjee2023benchmarkingllmpoweredchatbots}, leaf, text width = 20em
                          ]
                        ]
                        [
                            \citet{gusev2024pingpongbenchmarkroleplayinglanguage}{,} \citet{gritta2024humanrankevalautomaticevaluationlms}{,}\citet{zhou-etal-2023-simoap}{,} \\  
\citet{luo2022alichilargescalemultimodaldataset}{,} \citet{Mehnaz2021GupShupSO}{,} \\
\citet{komma2023accurategeneralizableevaluationmetrics}{,}\citet{budzianowski-etal-2018-multiwoz}{,} \\  
\citet{app10030762}{,} \citet{zhang2024probingmultiturnplanningcapabilities}{,} \\  
\citet{mehri2020dialogluenaturallanguageunderstanding}{,} \citet{zhang2018personalizingdialogueagentsi}, leaf, text width = 20em
                        ]
                    ]
                    [
                        Annotation As The Exact Result For Matching, text width = 20em
                        [
                            \citet{han-etal-2023-log}{,} \\  
\citet{xing2022evaluatingimprovingcontextattention}{,} \citet{10.1145/3545570}{,} \\  
\citet{yang2021ubarfullyendtoendtaskoriented}{,} \citet{dong2024bamboocomprehensivebenchmarkevaluating}{,} \citet{wu2024benchmarkinglargelanguagemodels}{,} \\  
\citet{budzianowski2020multiwozlargescalemultidomain}{,} \citet{byrne-etal-2019-taskmaster}, leaf, text width = 22em
                        ]
                    ]
                ]
                [
                  Annotation-free Evaluation
                  [
                    Point-wise Response Scoring, text width = 15em
                    [
                        \citet{jones-etal-2024-multi}{,} \citet{ferron-etal-2023-meep}{,} \citet{lin-chen-2023-llm}{,} \\  
\citet{fan-etal-2024-sedareval}{,} \citet{10.1007/978-3-662-49390-8_35}{,} \citet{knowledge2010004}{,} \\  
\citet{Zhang_2021}{,} \citet{li2024fbbenchfinegrainedmultitaskbenchmark}{,} \citet{a-2022-automating}{,} \\  
\citet{shi-etal-2023-rade}{,} \citet{ren2023cpmiconditionalpointwisemutual}{,} \citet{jia2023multiturnresponseselectionusing}, leaf, text width = 25em
                    ]
                  ]
                  [
                    Pair-wise or List-wise Response Scoring, text width = 20em
                    [
                        \citet{kim-etal-2024-debate}{,} \citet{ruiz-dolz-etal-2023-automatic}{,} \\  
\citet{liu-etal-2024-empirical}{,} \citet{DBLP:conf/nips/BaiY0LHWYZXLZLH23}{,} \\  
\citet{lee2020evaluationprotocolgenerativeconversational}{,} \citet{10.1214/aos/1079120141}{,} \citet{NIPS2006_f44ee263}{,} \\  
\citet{park2024pairevalopendomaindialogueevaluation}{,} \citet{li2019acuteevalimproveddialogueevaluation}{,} \\  
\citet{knowledge2010004}{,} \citet{banerjee-lavie-2005-meteor}{,} \\  
\citet{Robertson2004UnderstandingID}{,}\citet{chan2023chatevalbetterllmbasedevaluators}, leaf, text width = 22em
                    ]
                  ]
                ]
            ]
        \end{forest}
    }
    \caption{Taxonomy of Evaluation Metrics for LLM-based Agents in Multi-Turn Conversations.}
    \label{fig:evaluation_metrics}
    \vspace{-5mm}
\end{figure*}

%% file: evaluation_method_comparison.tex
\begin{table}[ht]
\centering
\caption{Comparison of Annotation-based vs.\ Annotation-free Method}
\label{tab:annotation-vs-llm-judge}
\begin{tabular}{@{}p{3cm} p{5.5cm} p{5.5cm}@{}}
\toprule
\textbf{Aspect} & \textbf{Annotation-based Evaluation} & \textbf{Annotation-free Evaluation} \\
\midrule

\textbf{Approach / Data} &
\begin{minipage}[t]{\linewidth}
\begin{itemize}[leftmargin=*]
  \item Relies on expert-annotated reference outputs
  \item Uses fixed labels (e.g.\ annotated intents, entities) for exact matching.
\end{itemize}
\end{minipage}
&
\begin{minipage}[t]{\linewidth}
\begin{itemize}[leftmargin=*]
  \item Uses model-based metrics (e.g.\ LLM scoring of relevance, coherence).
  \item Can perform direct, pairwise or listwise comparisons without gold labels.
\end{itemize}
\end{minipage}
\\
\midrule

\textbf{Advantages} &
\begin{minipage}[t]{\linewidth}
\begin{itemize}[leftmargin=*]
  \item \textbf{High Precision}: Ground-truth references ensure accurate scoring.
  \item \textbf{Interpretability}: Clear semantics of annotated labels guide evaluation.
\end{itemize}
\end{minipage}
&
\begin{minipage}[t]{\linewidth}
\begin{itemize}[leftmargin=*]
  \item \textbf{Flexibility}: Adaptable to open-ended qualities without new data.
  \item \textbf{Automation}: Easily applies to new domains without annotation overhead.
\end{itemize}
\end{minipage}
\\
\midrule

\textbf{Disadvantages} &
\begin{minipage}[t]{\linewidth}
\begin{itemize}[leftmargin=*]
  \item \textbf{Costly \& Slow}: Requires extensive human effort and time to produce ground truth label.
  \item \textbf{Bias}: Subject to annotator bias and low inter‐annotator agreement.
\end{itemize}
\end{minipage}
&
\begin{minipage}[t]{\linewidth}
\begin{itemize}[leftmargin=*]
  \item \textbf{Reliability \& Bias Issues}: LLM evaluations can be inconsistent and biased.
  \item \textbf{Per-Query Cost}: API latency or usage costs for large LLM calls.
\end{itemize}
\end{minipage}
\\
\midrule

\textbf{Flexibility \& Coverage} &
\begin{minipage}[t]{\linewidth}
\begin{itemize}[leftmargin=*]
  \item \textbf{Rigid}: Exact matching penalizes correct but unannotated responses.
  \item \textbf{Static}: Hard to cover new scenarios without fresh annotations.
\end{itemize}
\end{minipage}
&
\begin{minipage}[t]{\linewidth}
\begin{itemize}[leftmargin=*]
  \item \textbf{Adaptive}: Can evaluate novel or broad cases via prompt design.
  \item \textbf{Broad Scope}: Judges subjective and multi-dimensional aspects dynamically.
\end{itemize}
\end{minipage}
\\
\bottomrule
\end{tabular}
\end{table}

%% file: combined-tables.tex
\begin{table}[ht]
\centering
\caption{Benchmarks for Task Multitask Capabilities, Interaction Patterns, and Temporal Dimensions, Safety and Tool-Use in Multi-Turn Conversation Systems}
\label{tab:summary_table_1}
\resizebox{0.85\textwidth}{!}{%
\begin{tabular}{|p{3cm}|p{5.5cm}|p{5.5cm}|}
\hline
\textbf{Study} & \textbf{Focus} & \textbf{Metrics} \\
\hline
\multicolumn{3}{|c|}{\textbf{Multitask Capabilities}} \\
\hline
\citet{Hendrycks2020MeasuringMM} & Domain-Specific Multitasking & Accuracy in domain-specific question answering and problem-solving \\
\hline
\citet{li2017dailydialogmanuallylabelledmultiturn} & Manually Annotated Multi-Turn Data (DailyDialog) & Annotation quality, response coherence \\
\hline
\citet{gopalakrishnan2019topical} & Topic-Based Dialogue Dataset (Topical Chat) & Knowledge grounding, conversational diversity \\
\hline
\multicolumn{3}{|c|}{\textbf{Interaction Patterns in Multi-Turn Conversations}} \\
\hline
\citet{Zheng2023JudgingLW} & Multi-Turn Benchmark (MT-Bench) & Precision, quality of generated responses \\
\hline
\citet{kwan2024mtevalmultiturncapabilitiesevaluation} & Human-LLM Interaction Patterns & Recognition of multi-turn patterns, task success rates \\
\hline
\multicolumn{3}{|c|}{\textbf{Temporal Dimensions in Multi-Turn Conversations}} \\
\hline
\citet{xu2021goldfishmemorylongtermopendomain} & Long-Term Conversations Dataset & Retention of shared context over sessions \\
\hline
\citet{maharana2024evaluatinglongtermconversationalmemory} & Months-Long Conversations Dataset & Context retention in multi-month interactions \\
\hline
\multicolumn{3}{|c|}{\textbf{Safety in Multi-Turn Conversations}} \\
\hline
\citet{yu2024cosafeevaluatinglargelanguage} & Safety: Coreference-Based Attacks & Vulnerability to coreference-based attacks \\
\hline
\multicolumn{3}{|c|}{\textbf{Tool-Use Benchmarks}} \\
\hline
\citet{DBLP:conf/iclr/MialonF0LS24} & real word question that is easy for human but hard for AI (GAIA benchmark) & Performance is gauged by the percentage of exact-match. \\
\hline
\citet{huang2024metatoolbenchmarklargelanguage} & Tool Awareness \& Decision-Making & Correct tool selection rate, effective usage in single-/multi-tool tasks \\
\hline
\citet{wang2024mintevaluatingllmsmultiturn} & Multi-Turn Tool-Use in Python & Successful integration \& execution of Python-based tools \\
\hline
\citet{xu2023toolmanipulationcapabilityopensource} & API Call Execution (ToolBench) & Function call success rate \\
\hline
\citet{anonymous2024tooldial} & ToolDial benchmark & Annotation efficiency, action prediction \\
\hline
\citet{shen2023hugginggptsolvingaitasks} & Orchestrated Tool-Use (HuggingGPT) & Task execution accuracy \\
\hline
\citet{li-etal-2023-api} & API Benchmarking & Context retention, API success rate \\
\hline
\citet{wang2024mtubenchmultigranularitytoolusebenchmark} & Multi-Granularity Tool-Use (MTU-Bench) & Tool selection, parameter accuracy \\
\hline
\end{tabular}
}
\end{table}

\begin{table}[ht]
\centering
\caption{Benchmarks for Reliability in Tool-use, Next-turn Response, Query Rewritten, Conversation Memory, Complete Interaction, Memory Editing, Task Representation, Context Modeling, Task Decomposition and Plan Selection in Multi-Turn Conversation Systems}
\label{tab:summary_table_2}
\resizebox{0.85\textwidth}{!}{%
\begin{tabular}{|p{3cm}|p{5.5cm}|p{5.5cm}|}
\hline
\textbf{Study} & \textbf{Focus} & \textbf{Metrics} \\
\hline
\multicolumn{3}{|c|}{\textbf{Reliability and Hallucination Mitigation in Tool-Use}} \\
\hline
\citet{zhang2024toolbehonestmultilevelhallucinationdiagnostic} & Multi-Level Hallucination Diagnosis & Hallucination detection accuracy, depth and breadth analysis \\
\hline
\multicolumn{3}{|c|}{\textbf{Next-Turn Response}} \\
\hline
\citet{moniri2024evaluatingperformancelargelanguage} & Multi-Round Structured Debates & Clarity, factual accuracy, consistency \\
\hline
\citet{abercrombie-batista-navarro-2020-parlvote} & Parliamentary Debate Analysis & Logical flow, structured reasoning \\
\hline
\multicolumn{3}{|c|}{\textbf{Query Rewritten and Retrieved Items}} \\
\hline
\citet{Aliannejadi_2020} & Relevant Utterance Dataset & Precision, Recall and F1-score \\
\hline
\multicolumn{3}{|c|}{\textbf{Conversation Memory}} \\
\hline
\citet{chen-etal-2024-socialbench} & Conversational Memory Benchmark & Keyword recall, memory retention over multiple turns \\
\hline
\multicolumn{3}{|c|}{\textbf{Complete Interaction}} \\
\hline
\citet{li2023how} & Long-Range Context Retrieval & Long-context recall, retrieval accuracy \\
\hline
\citet{maharana2024evaluatinglongtermconversationalmemory} & Long-Term Multi-Agent Conversations & Memory retention, coherence in extended dialogues \\
\hline
\multicolumn{3}{|c|}{\textbf{Task Representation}} \\
\hline
\citet{zhang2024lammapgeneralizablemultiagentlonghorizon} & Language Model-Driven PDDL Planning & Planning efficiency, generalization across tasks \\
\hline
\multicolumn{3}{|c|}{\textbf{Context Modeling}} \\
\hline
\citet{xiao-etal-2024-flowbench} & Reducing Context Hallucinations & Hallucination reduction, response precision \\
\hline
\multicolumn{3}{|c|}{\textbf{Task Decomposition}} \\
\hline
\citet{wang2024tdagmultiagentframeworkbased} & Multi-Agent Task Decomposition & Adaptability, task efficiency \\
\hline
\citet{wu2024mldtmultileveldecompositioncomplex} & Multi-Level Task Planning & Task decomposition accuracy \\
\hline
\multicolumn{3}{|c|}{\textbf{Post-Generation Plan Selection}} \\
\hline
\citet{kargupta-etal-2024-instruct} & Structured Questioning & Plan consistency, dialogue coherence \\
\hline
\end{tabular}
}
\end{table}